\documentclass[10pt,twocolumn,letterpaper]{article}

\usepackage{3dv}
\usepackage{times}
\usepackage{epsfig}
\usepackage{graphicx}
\usepackage{amsmath}
\usepackage{amssymb}
\usepackage{color}
\usepackage{multirow}
\usepackage{authblk}
\usepackage{gensymb}
\usepackage[hyphens]{url}

\usepackage[pagebackref=true,breaklinks=true,letterpaper=true,colorlinks,bookmarks=false]{hyperref}

\threedvfinalcopy % *** Uncomment this line for the final submission

\def\threedvPaperID{194} % *** Enter the 3DV Paper ID here

\ifthreedvfinal\fi
\begin{document}

%%%%%%%%% TITLE
% \title{Learning Implicit 3D Representations of Dressed Humans from Sparse Views}
% \title{Data-driven 3D Reconstruction of Dressed Humans from Sparse Views}
\title{Data-Driven 3D Reconstruction of Dressed Humans From Sparse Views}

\author{
Pierre Zins$^{1,2}$~~~
Yuanlu Xu$^2$~~~
Edmond Boyer$^1$~~~
Stefanie Wuhrer$^1$~~~
Tony Tung$^2$~~~
% \smallskip 
\vspace{-3mm}
\\
$^1$Univ. Grenoble Alpes, Inria, CNRS, Grenoble INP \thanks{Institute of Engineering Univ. Grenoble Alpes}, LJK, 38000 Grenoble, France
\\
$^2$Facebook Reality Labs, Sausalito, USA
\\
\small{\textit{ name.surname@inria.fr, merayxu@gmail.com, tony.tung@fb.com}}
}
% \author[1]{Pierre Zins}
% \author[2]{Yuanlu Xu}
% \author[1]{Edmond Boyer}
% \author[1]{Stefanie Wuhrer}
% \author[2]{Tony Tung}
% \affil[1]{Univ. Grenoble Alpes, Inria, CNRS, Grenoble INP \thanks{Institute of Engineering Univ. Grenoble Alpes}, LJK, 38000 Grenoble, France}
% \affil[2]{Facebook Reality Labs, Sausalito, USA}
% \affil[ ]{\normalsize \textit {name.surname@inria.fr, merayxu@gmail.com, tony.tung@fb.com}}
% \affil[ ]{\normalsize \textit {pierre.zins@inria.fr, merayxu@gmail.com, edmond.boyer@inria.fr}}
% \affil[ ]{\normalsize \textit{stefanie.wuhrer@inria.fr, tony.tung@fb.com}}


\maketitle

\thispagestyle{empty}
%%%%%%%%% ABSTRACT
\begin{abstract}

Recently, data-driven single-view reconstruction methods have shown great progress in modeling 3D dressed humans. However, such methods suffer heavily from depth ambiguities and occlusions inherent to single view inputs.
In this paper, we tackle this problem by considering a small set of input views and investigate the best strategy to suitably exploit information from these views. We propose a data-driven end-to-end approach that reconstructs an implicit 3D representation of dressed humans from sparse camera views. Specifically, we introduce three key components: first a spatially consistent reconstruction that allows for arbitrary placement of the person in the input views using a perspective camera model; second an attention-based fusion layer that learns to aggregate visual information from several viewpoints; and third a mechanism that encodes local 3D patterns under the multi-view context. 
In the experiments, we show the proposed approach outperforms the state of the art on standard data both quantitatively and qualitatively. To demonstrate the spatially consistent reconstruction, we apply our approach to dynamic scenes. Additionally, we apply our method on real data acquired with a multi-camera platform and demonstrate our approach can obtain results comparable to multi-view stereo with dramatically less views. Code is released at \url{https://gitlab.inria.fr/pzins/data-driven-3d-reconstruction-of-dressed-humans-from-sparse-views/}.

\end{abstract}

%%%%%%%%% BODY TEXT
\vspace{-2mm}
\section{Introduction}

The ability to produce accurate visual models of real humans in every-day context, in particular with their clothing and accessories, is useful in a wide range of applications that deal with captured human avatars, typically in the virtual and augmented reality or telepresence domains. Using images for that purpose has been an active field of research for decades, with issues that result, in part, from the high dimensionality of the space of human shapes and appearances, especially with dressed people. The challenge is accentuated when only few viewpoints are considered, a situation that is, on the other hand, common in many practical contexts, for instance with mobile devices. While model based strategies (\eg, SMPL~\cite{DBLP:journals/tog/LoperM0PB15}) have shown impressive results in case of undressed bodies, they cannot easily generalize to generic humans with clothing and accessories. This paper investigates how to recover such 3D models by combining information from sparse calibrated views. 

\begin{figure}[t!]
\begin{center}
\includegraphics[width=0.87\linewidth]{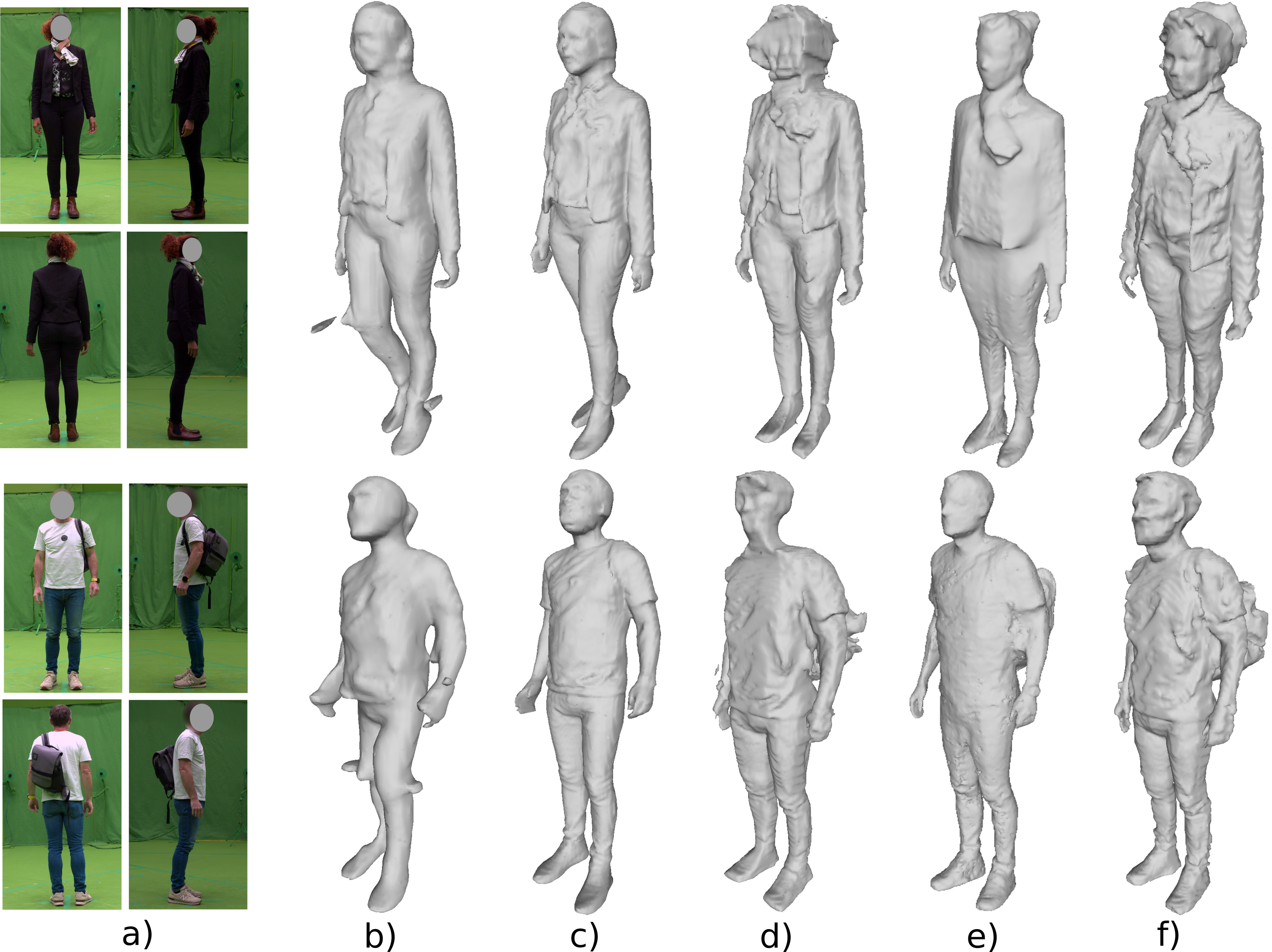}
\end{center}
\vspace{-3mm}
\caption{a) Real scene cropped images. b) PIFu~\cite{pifuSHNMKL19} and c) \mbox{PIFuHD}~\cite{DBLP:conf/cvpr/SaitoSSJ20} with a single frontal view. d) PIFu with $4$ views. e) Multi-view stereo~\cite{DBLP:conf/eccv/LeroyFB18} reconstruction with 60 views. f) Our method with $4$ views.}
\label{fig:teaser}
\vspace{-5mm}
\end{figure}

Acquiring 3D human models from images is a long standing research topic in computer vision. When images from several viewpoints are available, multi-view stereo approaches (\eg \cite{DBLP:conf/cvpr/SeitzCDSS06,DBLP:conf/cvpr/FurukawaP07}), and their learning based extensions (\eg \cite{ji2017surfacenet,DBLP:conf/eccv/LeroyFB18}), allow for highly detailed 3D reconstructions by combining multi-view information with photo-consistency criteria. This generative strategy builds on photo-metric redundancy among input images and tends to fail with only sparse viewpoints. Besides, data-driven reconstruction methods, that only require a single view, have been proposed. This includes methods based on low-dimensional parametric models (\eg \cite{DBLP:conf/cvpr/PavlakosCGBOTB19})  which are anyway limited with clothing and accessories; methods based on volumetric representations (\eg \cite{DBLP:conf/eccv/VarolCRYYLS18}) with bounded level-of-details by construction; and methods based on implicitly defined continuous neural representations (\eg \cite{pifuSHNMKL19}). These latter methods have demonstrated their ability to recover humans with clothing and accessories. Yet, the single-view reconstruction problem is highly ambiguous and results easily suffer from artifacts when the input scene differs substantially from the training set. To remedy this, methods accounting for multiple input views have been proposed~\eg \cite{DBLP:conf/eccv/HuangLCZXLLML18,pifuSHNMKL19}. These extensions, however, merely combine single-view estimations with simple average pooling. Such ways of fusion do not fully exploit multi-view cues and are still plagued by single-view ambiguities.

In this paper, we adopt the widely approved implicit neural representations and focus on multi-view fusion. With respect to single-view estimation this task raises several issues. 
First, single-view reconstruction methods generally assume a person centered and scaled input image. This needs to be compensated for when dealing with sequences of moving humans and in order to obtain spatially consistent reconstruction with coherent localization and scales among the sequence frames.
The second question is how to aggregate local information from viewpoints that can differ significantly, for instance front and side-views, and which can therefore predict different occupancy at a given spatial location. The third issue is how to account for local contexts, defined by image color cues around a 3D point, that gain in variability with increasing views but also allow to better differentiate local geometric patterns. 
To address these issues, we propose a data-driven end-to-end approach that reconstructs a 3D model of the dressed human from sparse camera views using an implicit representation. Specifically, our method has three key components:

{\small \textbullet} \,A spatially consistent 3D reconstruction framework that allows for arbitrary placement of the human in the scene that uses the perspective camera model, achieved by learning the model in a canonical coordinate system and by accounting for the transformation of each input view to this system.

{\small \textbullet} \,A learnable attention-based fusion layer that weighs view contributions. This layer implements a multi-head self-attention mechanism inspired by the transformer network \cite{DBLP:conf/nips/VaswaniSPUJGKP17}.

{\small \textbullet} \,A local 3D context encoding layer that better generalizes over the local geometric configurations, which is implemented through randomized 3D local grids.

In the experiments, we evaluate our approach against the state of the art on public benchmarks. To demonstrate the value of the spatially consistent reconstruction, we apply our method to dynamic scenes with large displacements. Moreover we also contribute with results on new real data obtained with a multi-view platform. They demonstrate the feasibility of data-driven approaches in practical real-world capture scenarios, even trained solely on synthetic data. 

\section{Related Work}

In this section, we focus on methods that reconstruct the 3D geometry of humans, possibly in clothing. 

\textbf{Monocular 3D reconstruction}
is an ill-posed problem, as a result of depth ambiguities and occlusions. Dimension reduction with parametric models is a strategy that has been extensively studied in the past two decades. Early achievements use a set of simple geometric primitives to track and reconstruct humans from monocular video \eg \cite{DBLP:journals/cviu/PlankersF01,DBLP:journals/ijrr/SminchisescuT03}. Statistical human body models learned from 3D scans allow to infer the naked body shape from monocular depth images~\cite{DBLP:journals/tog/AnguelovSKTRD05} or color images~\cite{DBLP:conf/eccv/BalanB08,5539853,DBLP:conf/iccv/GuanWBB09,DBLP:conf/eccv/BogoKLG0B16,DBLP:conf/cvpr/KanazawaBJM18,DBLP:conf/cvpr/PavlakosZZD18,DBLP:conf/cvpr/KolotourosPD19,DBLP:conf/iccv/XuZT19}. Some of these models are even sufficiently detailed to allow capturing facial expressions and hand gestures~\cite{DBLP:conf/cvpr/PavlakosCGBOTB19}. More recent techniques tend to directly regress parameters of human body models with deep neural networks~\cite{DBLP:conf/cvpr/KanazawaBJM18,DBLP:conf/cvpr/PavlakosZZD18,DBLP:conf/iccv/XuZT19}. Another line of work uses a 3D template mesh as input and trains a deep neural network to deform or regress the template vertices given a monocular image~\cite{DBLP:conf/cvpr/PavlakosCGBOTB19,DBLP:conf/cvpr/0004ZWCY19}. All of these methods are limited to undressed human bodies, and cannot reconstruct clothing or accessories. Some methods allow nevertheless for clothing as offsets from a parametric body model based on an input monocular video~\cite{DBLP:conf/cvpr/AlldieckMXTP18,DBLP:conf/iccv/BhatnagarTTP19} or single image~\cite{DBLP:conf/iccv/ZhengYWDL19}, or using physics-based simulation~\cite{DBLP:conf/cvpr/YuZZZDPL19,DBLP:conf/cvpr/PatelLP20}. Using parametric models, some approaches allow for real-time reconstruction of dynamic humans from a single depth camera~\cite{DBLP:conf/iccv/BogoBL015,DBLP:conf/iccv/YuGXDSZLDL17,DBLP:journals/pami/YeSDPY16,DBLP:journals/pami/YuZZGDLPL20}. While parametric models allow for interesting solutions, the level of detail and variability of the reconstructed clothing and accessories remain inherently limited.
To overcome this problem, alternative representations have been explored. Volumetric representations~\cite{DBLP:conf/eccv/VarolCRYYLS18,DBLP:conf/eccv/JacksonMT18} and methods that estimate novel silhouettes to enable visual hull reconstruction~\cite{DBLP:conf/cvpr/NatsumeSH0MLM19} offer the advantage of allowing for more clothing variety at the cost of requiring large memory. Methods that represent the reconstruction using few depth maps~\cite{DBLP:conf/iccv/GabeurFMSR19,DBLP:conf/iccv/SmithLHM019} are less memory demanding, but cannot represent arbitrary clothing topology.
Many recent approaches address the problems of memory efficiency and resolution with implicitly-defined continuous neural representations. A seminal work that uses this representation to reconstruct humans from monocular images is PIFu~\cite{pifuSHNMKL19}, which learns pixel-aligned implicit functions to locally align image pixels with the global location of the 3D human. Follow-up methods increase the image resolution for higher levels of detail~\cite{DBLP:conf/cvpr/SaitoSSJ20}, propose animatable reconstructions~\cite{DBLP:conf/cvpr/HuangXL0T20}, combine PIFu with a volumetric representation or voxelized model to incorporate global 3D information~\cite{DBLP:conf/nips/HeCJS20,DBLP:journals/corr/abs-2007-03858}, and combine PIFu with a parametric model to allow for coherent body reconstruction~\cite{DBLP:conf/eccv/BhatnagarSTP20}. Alternative representations propose using tetrahedral truncated signed distance functions~\cite{DBLP:conf/cvpr/OnizukaHTSUT20}, and using periodic activation functions to better capture high frequencies~\cite{DBLP:conf/nips/SitzmannMBLW20}. 

Our work also builds on implicit neural representations for their ability to efficiently encode shape information. However, departing from the single view paradigm we focus on how to leverage several views to overcome some of the limitations of single-view inference.

\textbf{Multiple View Reconstruction}
has been researched extensively, and a full review is beyond the scope of this paper. Classical stereo and multi-view stereo techniques reconstruct 3D geometry from a set of images under assumptions, especially photo-coherent Lambertian surfaces, \eg, \cite{DBLP:conf/cvpr/SeitzCDSS06,DBLP:conf/cvpr/FurukawaP07}. More recent methods allow for improved results by learning some parts of the classical multi-view stereo pipelines like the photo-consistency~\cite{DBLP:conf/eccv/LeroyFB18} or the depth maps fusion~\cite{DBLP:conf/cvpr/DonneG19}. These methods require short baselines and many views~\cite{DBLP:conf/iccv/FrancoB05,DBLP:journals/ijcv/KutulakosS00}, which lead to practical limitations.
Methods based on Neural Radiance Field (NeRF) achieve photo-realistic rendering but also require numerous views or images (typically more than 50, up to hundreds) for learning an MLP that represents the scene. They are usually scene specific although some limitations have been explored in recent work~\cite{DBLP:conf/eccv/MildenhallSTBRN20,DBLP:journals/corr/abs-2012-02190,DBLP:conf/nips/SitzmannZW19,DBLP:journals/corr/abs-2012-15838}.

When focusing on humans, several previous methods use a template-based approach to reconstruct the 3D geometry from silhouette information~\cite{DBLP:journals/tog/AguiarSTAST08,DBLP:journals/tog/VlasicBMP08,DBLP:conf/cvpr/GallSATRS09,DBLP:conf/3Dim/DibraJOZG16}. Some techniques take advantage of low-dimensional parametric models to reconstruct the 3D body from multiple RGB images~\cite{DBLP:conf/cvpr/BalanSBDH07,DBLP:conf/cvpr/JooSS18}. When multiple depth images are available, a full 3D human model can be reconstructed by fitting a parametric model to the scans~\cite{weiss-2011} or by globally registering and merging the depth images~\cite{DBLP:journals/sigpro/LiuQBYHTH15}.

Closer to our work, two methods~\cite{DBLP:conf/eccv/HuangLCZXLLML18,pifuSHNMKL19} propose the use of implicit representations to reconstruct 3D humans from multiple input images. Unlike our work, these methods combine the views by simply
averaging their contributions. They do not handle visibility consistency among the different views, and in particular occlusions. In this paper, we propose a solution based on an implicit representation with a novel learnable attention-based fusion layer that efficiently weighs the available views and outputs a spatially consistent reconstruction.

%%%%%%%%%%%%%%%%%%%%%%%%%%%%
\section{Method}

In this section we first give an overview of our method and explain the representation that is used. We then present our strategy to learn and infer humans in a large scene and our contributions with the spatially consistent reconstruction, the attention-based fusion layer and the local context learning. 

%%%%%%%%%%%%%%%%%%%%%%%%%%%%
\subsection{Overview}

\begin{figure*}[t]
\begin{center}
\includegraphics[width=1.0\linewidth]{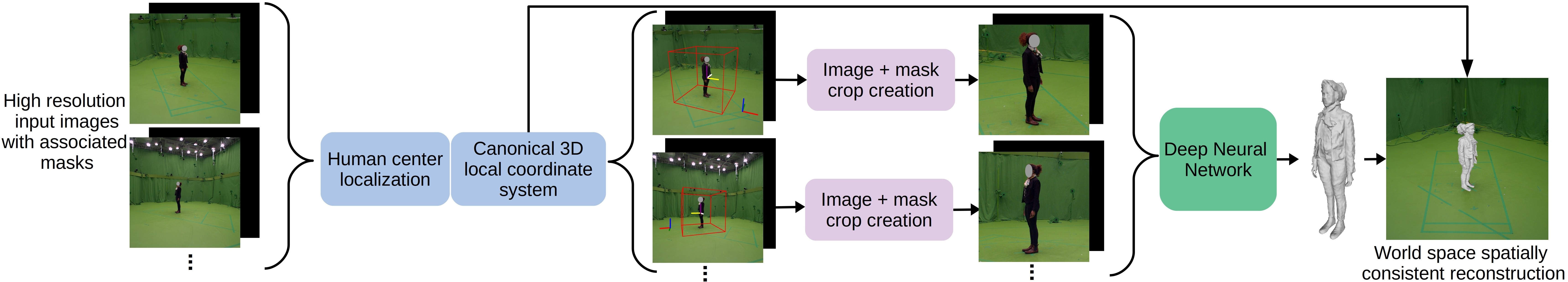}
\end{center}
\vspace{-3mm}
\caption{Overview of the proposed pipeline. Given a sparse set of input images with associated background masks and known calibration, our method reconstructs a spatially consistent 3D model.}
\label{fig:overview}
\vspace{-3mm}
\end{figure*}

\begin{figure*}[t]
\begin{center}
\includegraphics[width=1.0\linewidth]{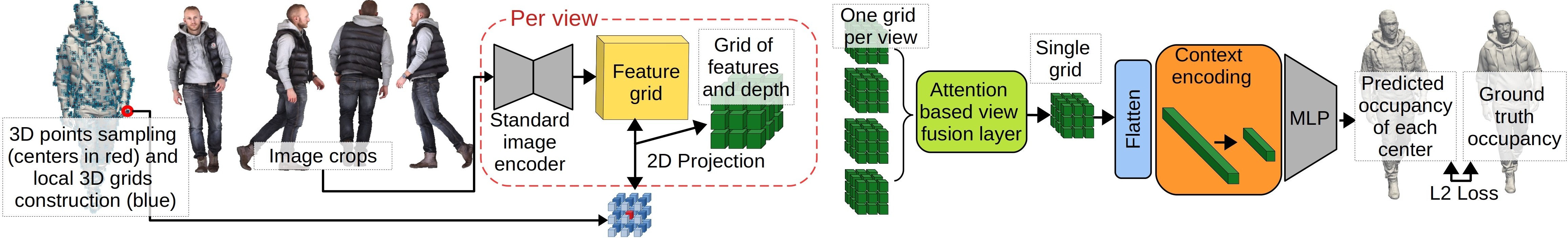}
\end{center}
\vspace{-3mm}
\caption{Overview of the deep neural network for multi-view training. Image features are extracted per view, and queried for a local grid around each sample. All views are integrated using an attention-based fusion layer, and a context encoding layer based on 3D convolution is applied before predicting occupancy.}
\label{fig:training}
\vspace{-3mm}
\end{figure*}

Our pipeline is described in Fig.~\ref{fig:overview}. High resolution images of a human and background masks are used as inputs to reconstruct a spatially consistent 3D model using an implicit representation. To allow for a spatially consistent reconstruction with proper scales and localization, we learn the model in a canonical 3D local coordinate system, and transform each observation to this space. This is achieved by localizing the 2D center of the human in each view, by triangulating to find the 3D position of the human center, and by defining a canonical 3D local coordinate system based on this information. This allows to create canonical crops of the input images and background masks so they can be fed to our deep neural network that learns to predict an implicit 3D reconstruction in a canonical space. The result, combined with the canonical 3D local coordinate system, allows to reconstruct a spatially consistent 3D model in the scene by placing the reconstruction in world coordinates. 

Fig~\ref{fig:training} gives an overview of our deep neural network for multi-view 3D reconstruction. Image features are first extracted using a standard multi-scale image encoder. Please refer to the supplementary materials for an ablation study on the image encoder. We then sample points by combining two strategies: random sampling in a 3D bounding box and importance sampling close to the surface with half of the points inside the mesh and the other half outside. We also construct a local 3D grid around each sample. Here we describe the method for a single sample but in practice a large number of points are processed in parallel. Using projection and bilinear interpolation, each point of the local grid is associated with a 2D feature, which is concatenated with the depth of the point. It is important to note that the previous steps are performed per-view and in the end a 3D local grid of features is obtained for each view. An attention-based module efficiently combines the information from the different views by merging the 3D local grids. A second fully connected fusion layer extracts a final 3D feature from the local grid. At inference time, we define a grid at the desired resolution, evaluate the occupancy function at every grid location, and apply the Marching Cubes algorithm~\cite{lorensen-1987} with a pre-defined threshold of 0.5 to recover a 3D mesh.

%%%%%%%%%%%%%%%%%%%%%%%%%%%%
% \subsection{Implicit 3D surface representation}
% \subsection{Person-centric Depth Origin}
\subsection{Multi-view Implicit Surface Representation}
\label{sec:implicit_3d_surface_representation}

Following recent progresses in learning-based shape modeling, we use an implicit 3D surface representation for the reconstruction task. Implicit surface representation converts arbitrary mesh surfaces into a function defined on a volume and allows for geometric details to be represented at arbitrary resolution. Furthermore, the use of neural implicit representations is memory-efficient and solves the main issue in other volumetric representations. Similar to methods like~\cite{DBLP:conf/cvpr/SaitoSSJ20,pifuSHNMKL19}, our implicit function takes the combination of pixel-aligned features with depth values as input and predicts an occupancy probability $o \in [0, 1]$.

When reconstructing from a single image, the conditioning on the depth is necessary to differentiate points on the same camera ray as their appearance features are the same. In our case with multiple views, associations of features can discriminate points of the same view line but the conditioning on the depth is still helpful to capture details as the spatial resolution of the features is limited. To optimally benefit from this conditioning, training examples should all be aligned so that the network can learn a prior of the depth from the training set. Therefore even if we consider reconstruction in large scene, we work in a canonical local coordinate system during training and at inference. The origin of the coordinate system is defined at the center of each training mesh and its orientation is the same as the world coordinate system, so we have the following equation: 
$ X_{local}^j = X_{world} + T_j $, 
where $T_j$ is the translation between the world origin and the center of the $j$-th mesh.
The exact definition of the center of a mesh is arbitrary but should be consistent for all the training examples. In practice, we use the median over all mesh vertices  for the $x$ and $z$ coordinates and the mean between the highest and the lowest vertices for the vertical coordinate $y$.
For each 3D point, the depth value given as input of the implicit function is its $z$-coordinate in the local coordinate system aligned with each of the cameras by applying the rotation $R_i$.
The implicit function takes the form:
% \vspace{-5mm}
\begin{equation}
\begin{aligned}
& f(E_I(K_i \left[ R | t \right]_i X_w), z(R_i X_{local}); \theta) = o,\\
& \left[\left|E\right| \times \mathbb{R}\right] \mapsto [0,1] &
\end{aligned}
% \vspace{-5mm}
\end{equation}
where $X_w$ is the 3D point in world coordinates, $K_i$ and $\left[ R | t \right]_i$ are respectively the intrinsic and extrinsic parameters of the $i$-th camera, $o$ is the occupancy probability at $X_w$, and $\left|E\right|$ the dimension of the 2D image feature. $E_I(...)$ is defined at any location in the image using bilinear interpolation of the values of $E_I$ at pixel locations.

%\subsection{Reconstruction in a large scene}
\subsection{Spatially Consistent Reconstruction}
Most existing works based on pixel-aligned features and implicit representation consider orthographic projection where the appearance of a subject is the same at any position in the scene. In single-view reconstruction, this simplified scenario removes the ambiguity between the size of the subject and its distance from the camera. On the contrary we deal with perspective projection like in real environments with the pinhole camera model. We consider the case where enough views are available to avoid the size versus distance ambiguity. To accommodate for perspective deformations, we augment the data during training by randomly placing the subjects in the scene.
As we are learning an implicit representation in a canonical 3D local coordinate system, the reconstruction at inference is inconsistent with the world space. Previous work tackles this problem with a neural network that estimates the spatial transformation of humans from a single image~\cite{DBLP:journals/corr/abs-2104-09283}. In our context, we propose to take advantage of the multiple views and triangulate the 3D coordinates from multiple 2D detections of the center of the human as shown in Fig.~\ref{fig:overview}. The 2D center positions are known at training time and predicted during inference using a convolutional deep neural network. The exact definition of the center of a human should be coherent with the point used to define the origin of the canonical coordinate systems. To supervise this network, we can use a similar dataset as in the remaining pipeline. Knowing the 3D center position, we can define a canonical 3D local coordinate system, perform the inference in that space and replace the result in world coordinates.
Note that the height of the subject is preserved as we do not apply any normalization on the size of the meshes during training.

%%%%%%%%%%%%%%%%%%%%%%%%%%%%
\subsection{Attention-based Fusion Layer}

Image-based reconstruction benefits from multi-view cues, \eg, stereo vision, which should be combined before the reconstruction is carried out in order to avoid premature single-view decisions and therefore limit ambiguities. Each view provides a feature and the question is how to aggregate them. Concatenating all the features, while simple, does not appear optimal because the fused features may become large when many images are considered, making it impossible to learn from an arbitrary numbers of views. Concatenation also imposes an order between views, which is undesirable in practice. 

Besides concatenation, fusion approaches based on statistics, such as sum-pooling~\cite{Eslami1204}, average pooling~\cite{DBLP:journals/corr/abs-1709-03019} or max pooling~\cite{DBLP:conf/iccv/SuMKL15} were proposed in the literature. 
The advantages are simplicity and invariance to both the order and the number of views. However, pooling operation loses information about individual view contributions. In particular, views in which a point is visible are considered equal to views in which the point is occluded and, more generally, erroneous information from an input view will contaminate the final prediction.

We propose to go one step further by learning the fusion and contextualising the information from different views. Previous work~\cite{DBLP:journals/ijcv/YangWMT20} proposes a simple learned fusion layer that computes a normalized score for each view, for each channel of a global feature. 
% Combined with a specific training strategy, they show good results on multi-view object reconstruction. 
The main limitation is that the score of each view is computed individually without taking into account the information from the other views.

\begin{figure}[t]
\begin{center}
\includegraphics[width=1.0\linewidth]{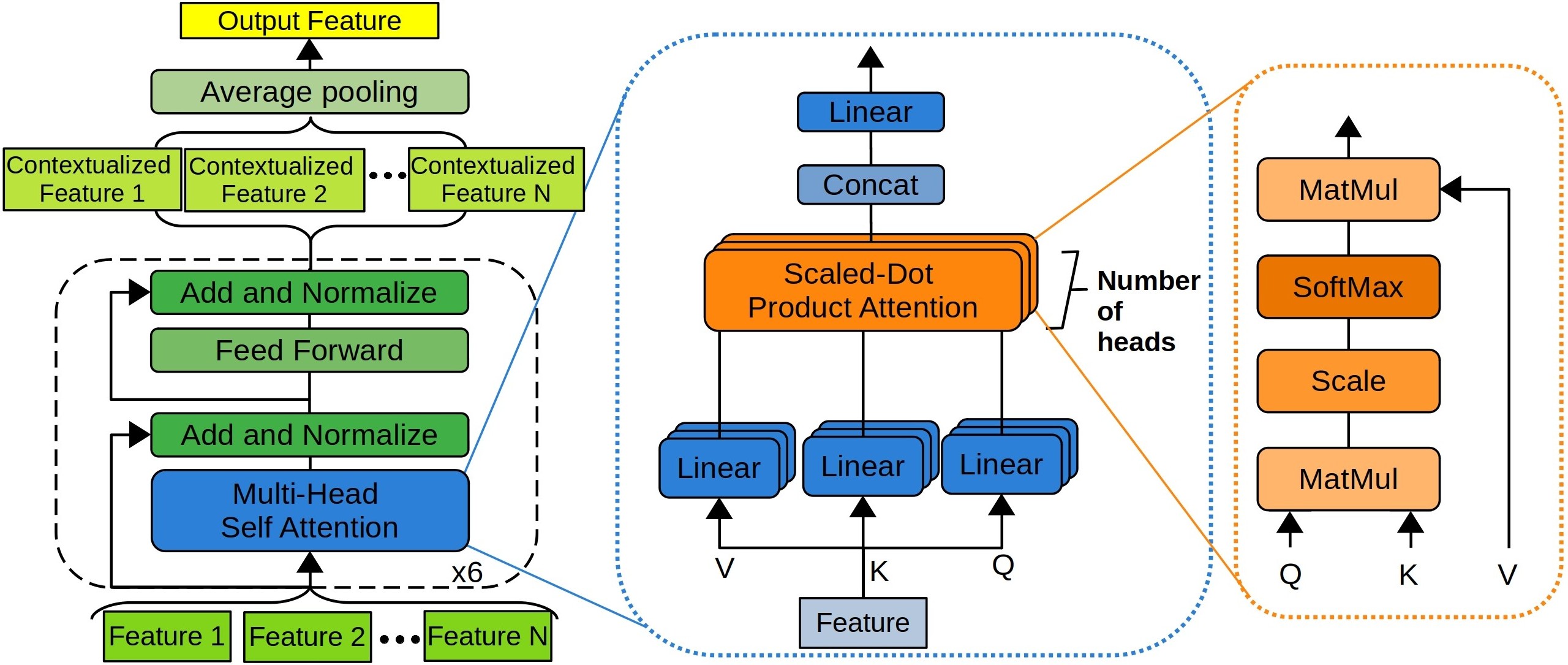}
\end{center}
\vspace{-3mm}
\caption{(Left) Our view fusion module. (Middle) Multi-Head Attention module. (Right) Scaled Dot-Product Attention.}
\label{fig:attention}
\vspace{-3mm}
\end{figure}

Inspired by recent progress in natural language processing to learn from sequences, we propose an architecture based on the transformer network~\cite{DBLP:conf/nips/VaswaniSPUJGKP17} which implements a multi-head self-attention mechanism and is described in Fig.~\ref{fig:attention}.
One key component is the \emph{scaled dot-product attention} which is a mapping function from a query along with a key / value pair to an output. The three vectors query $Q=M^qX$, key $K=M^kX$, and value $V=M^vX$ are the embedding of the original feature $X$ parameterized by matrices $M^q$, $M^k$ and $M^v$, respectively. The idea is to compute an attention score for each view based on a compatibility of a query with a corresponding key: 
%\vspace{-3mm}
\begin{equation}
    \text{Attention}(Q, K, V) = \mathbf{softmax}\left(\frac{Q K^T}{\sqrt{d_k}}\right) V,
%\vspace{-1mm}
\end{equation}
where $d_k$ the common dimension of $K$, $Q$ and $V$.

% The second main concept is to use multiple heads to allow the network to attend to different geometric patterns.
To allow the network to attend to different geometric patterns, we propose to use multiple heads. For that, $Q$, $K$ and $V$ are linearly projected $h$ times and processed in parallel through a scaled dot-product attention layer.
% with the learned parameters $W_i^q$, $W_i^k$ and $W_i^v$. Each projection of $Q$, $K$, $V$ is processed in parallel through a scaled dot-product attention layer. 
The results from the different heads are concatenated and finally projected once again to obtain the final output :
\vspace{-2mm}
\begin{equation}
\begin{aligned}
    &\text{MultiHead}(Q, K, V) = \mathbf{concat}(H_{1}, ..., H_{h}) W^{o}\\
    &\text{with } H_{i} = \text{Attention}(QW^{q}_{i}, KW^{k}_{i}, VW^{v}_{i})
\end{aligned}
\vspace{-2mm}
\end{equation}
where $W^{q}_{i}$, $W^{k}_{i}$, $W^{v}_{i}$ are respectively the parameters of the linear mapping of $Q$, $K$ and $V$, and $W^o$ the parameters of the final projection.

The output of the attention modules is a set of features. Each of them contains the original information from the corresponding view that now takes into account the information from all the other available views. Finally we use the mean of these features as output of our view fusion module. Note also, that we do not use any positional encoding on the input feature sequence to remain invariant to the view order.
% Two options exist to implement multi-head self-attention. The standard narrow version splits $Q$, $K$ and $V$ into small chunks and each head processes one of them. On the opposite, the wide option propagates entirely $Q$, $K$ and $V$ to each head. This version provides superior performance at the expense of computation time and memory requirements. In our work, we choose the narrow option which offers a very good compromise.

%%%%%%%%%%%%%%%%%%%%%%%%%%%%
\subsection{Local 3D Context Encoding}

\begin{figure}[t]
\begin{center}
\includegraphics[width=0.75\linewidth]{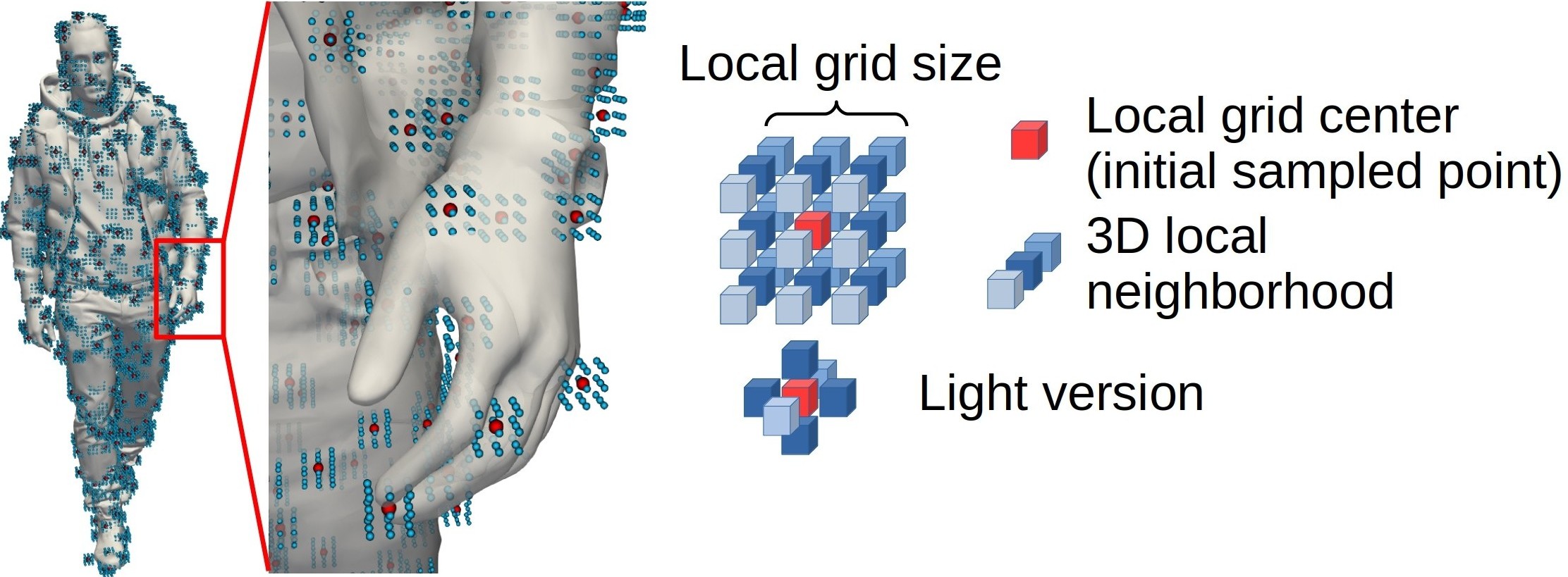}
\end{center}
\vspace{-3mm}
\caption{A local 3D grid is constructed around each sampled point (in red), and parameterized by a size and an orientation.}
\label{fig:local_grid}
\vspace{-3mm}
\end{figure}

In the proposed framework, projection is used to associate 3D points with 2D image features for each available view. Then, the attention-based fusion layer weighs the contribution of each view in the fused feature. Finally, a Multi-Layer Perceptron (MLP) predicts an occupancy probability. However, such features do not take the 3D geometric context into consideration since the neighbourhood is only considered in 2D when features are extracted from the images. To include 3D context, we propose to build a local 3D grid around each sampled point and associate each point of the local 3D grid with 2D image features by projection.

The attention-based layer is applied individually on each point of the local grids, after which we add another context fusion module that combines the information coming from a 3D neighbourhood of a sampled point. This module is shown in orange in Fig.~\ref{fig:training}, and is implemented with a fully connected layer. Thanks to this additional layer, the neural network is aware of the local 3D context of a point. In this way, we expect the network to better capture 3D geometric patterns and to increase robustness against nuisance factors (\eg, texture, lighting).

As shown in Fig.~\ref{fig:local_grid}, the local grid is parameterized by the size $S$ and orientation $R$.
Empirically, we found that fixing $R$ during training strongly links the local grid to the global coordinate system and the orientation of the human body. To remain invariant to the orientation of the human, during training we randomly align $R$ with one of the available views at each iteration.

The grid size $S$ needs to be chosen based on the training data and the type of the targeted 3D patterns. Our goal is to learn local 3D patterns that typically contain points in the same or close-by body parts. As a full local grid can be expensive in computation time and memory, we propose a variant that uses only the cells along the three grid axes that traverse the center of the grid. In that case, three one-dimensional vectors are considered instead of one three-dimensional grid, which significantly decreases the number of grid points while still allowing to take into account local context along three directions. We call this version "light" and use it in all our experiments. We also provide an ablation study on the grid size in supplementary materials.

%%%%%%%%%%%%%%%%%%%%%%%%%%%%
\section{Experimental Results}

In this section, we evaluate the proposed method and compare it with the state of the art. First, we give implementation details and introduce the training and testing datasets as well as the evaluation metrics. We then compare our approach quantitatively and qualitatively against the current state of the art and provide an ablation study to justify our contributions. Finally we show results of spatially consistent reconstruction and applications on real multi-view stereo data. Please refer to the supplementary materials for additional visual results and comparisons. 
%\textit{Code will be publicly available once this paper gets accepted}.
%  data which provide a good insight into the method's applicability

%%%%%%%%%%%%%%%%%%%%%%%%%%%%
\subsection{Implementation Details}

% We implemented our method using Pytorch~\cite{DBLP:conf/nips/PaszkeGMLBCKLGA19}.
Our human center localization network is implemented with the standard VGG16~\cite{DBLP:journals/corr/SimonyanZ14a} architecture. The image encoder or our reconstruction network is a Stacked Hourglass Network, with intermediate supervision, composed of 4 hourglass modules each of depth $2$. The size of the output features is $128\times 128\times 256$. Since we trained the network with a small batch size, we also introduced group normalization instead of batch normalization. Our view fusion layer is composed of 6 modules based on multi-head self-attention with 6 heads. The local 3D context fusion maps features from a $3\times 3\times 3$ grid into a single feature of size $256$. The Multi Layer Perceptron (MLP) is composed of 6 layers of dimensions $256, 1024, 512, 256, 128, 1$ with skip connections between the first layer and all the other layers except the last one. We optimized our network during $100$ epochs using the root mean square propagation algorithm with a learning rate of $1\times 10^{-4}$ that is divided by 10 at iterations 60 and 80. More details are available in the supplementary materials.

%%%%%%%%%%%%%%%%%%%%%%%%%%%%
\subsection{Settings}

% We adopt the settings from~\cite{pifuSHNMKL19,DBLP:conf/cvpr/SaitoSSJ20,DBLP:conf/cvpr/HuangXL0T20} and use data from Renderpeople~\cite{renderpeople} as our training dataset. 
We create our synthetic dataset with Renderpeople~\cite{renderpeople}, a public commercial dataset that provides highly detailed meshes obtained from 3D scans and corrected by artists. Its main advantage is the very high quality of the geometry which is essential to learn geometric details, especially with clothing. The humans from this set are in relatively standard poses and often hold accessories such as bags, cups or other objects. In total we have $1026$ meshes, split into $800$ meshes for training, $100$ for validation and $126$ for testing.

To evaluate quantitatively the reconstructed human meshes, we first compute the Chamfer Distance (\textbf{CD}) between the ground truth mesh and the reconstructed mesh. By considering average distances between meshes, this metric tends to measure the global quality of the reconstructions. To focus more on local details, we also consider surface normal of the reconstructed and ground truth meshes and compute the $\mathbb{L}_2$ and cosine distances between them (\textbf{Norm Cosine} and \textbf{Norm L2}, respectively). Finally, in order to evaluate accurately the raw predictions of our network before the Marching Cubes post-processing that transforms the occupancy probability grid into a mesh, we compute the average $\mathbb{L}_1$ distance ($\times 10^3$) between predicted and ground truth occupancy (\textbf{Occ L1}).

\begin{table}
\setlength{\tabcolsep}{3pt}
\resizebox{\linewidth}{!}{
\centering
\begin{tabular}{@{}l|c|c|c|c|c|c|c|c@{}}
    \hline
    \multicolumn{1}{c|}{\multirow{2}{*}{\textbf{Methods}}}  & \multicolumn{2}{c|}{\textbf{CD (cm) $\downarrow$}} & \multicolumn{2}{c|}{\textbf{Occ L1} $\downarrow$} & \multicolumn{2}{c|}{\textbf{Norm Cosine} $\downarrow$} & \multicolumn{2}{c}{\textbf{Norm L2} $\downarrow$}\\
    \cline{2-9}
    & mean & median & mean & median & mean & median & mean & median\\
    \hline
    PaMIR~\cite{DBLP:journals/corr/abs-2007-03858} & 0.554 & 0.508 & 1.977 & 1.754 & 0.097  & 0.090 & 0.361 & 0.343\\
    PIFu~\cite{pifuSHNMKL19} & 0.592 & 0.510 & 2.079 & 1.773 & 0.103  & 0.093 & 0.376 & 0.358\\
    PIFuHD~\cite{DBLP:conf/cvpr/SaitoSSJ20} & 2.008 & 1.624 & 5.837 & 4.543 & 0.181 & 0.162 & 0.544 & 0.503\\
    Ours & \textbf{0.367} & \textbf{0.316} & \textbf{1.538} & \textbf{1.323} & \textbf{0.089} & \textbf{0.083} & \textbf{0.350} & \textbf{0.337}\\
    \hline
\end{tabular}
}
% \vspace{0.5mm}
\caption{Quantitative results and comparisons with PaMIR~\cite{DBLP:journals/corr/abs-2007-03858}, PIFu~\cite{pifuSHNMKL19} and PIFuHD~\cite{DBLP:conf/cvpr/SaitoSSJ20} on Renderpeople dataset. PaMIR, PIFu and ours use 4 views as input (see Fig.~\ref{fig:sota}) and PIFuHD uses a single frontal view. Best scores are in \bf{bold}.}
\vspace{-2mm}
\label{tab:state_of_the_art}
\end{table}

\begin{table}
\setlength{\tabcolsep}{3pt}
\resizebox{\linewidth}{!}{
\centering
\begin{tabular}{@{}l|c|c|c|c|c|c|c|c@{}}
    \hline
    \multicolumn{1}{c|}{\multirow{2}{*}{\textbf{Variants}}}  & \multicolumn{2}{c|}{\textbf{CD (cm) $\downarrow$}} & \multicolumn{2}{c|}{\textbf{Occ L1} $\downarrow$} & \multicolumn{2}{c|}{\textbf{Norm Cosine} $\downarrow$} & \multicolumn{2}{c}{\textbf{Norm L2} $\downarrow$}\\
    \cline{2-9}
    & mean & median & mean & median & mean & median & mean & median\\
    \hline
    w./o. fusion & 0.553 & 0.478 & 2.013 & 1.755 & 0.101 & 0.093 & 0.373 & 0.353\\
    w./o. context & 0.413 & 0.363 & 1.622 & 1.399 & 0.091 & 0.087 & 0.353 & 0.342\\
    Ours full & \textbf{0.367} & \textbf{0.316} & \textbf{1.538} & \textbf{1.323} & \textbf{0.089} & \textbf{0.083} & \textbf{0.350} & \textbf{0.337}\\ 
    \hline
\end{tabular}
}
% \vspace{0.5mm}
\caption{Ablation studies on the effectiveness of different components. We evaluate our method when deactivating the view fusion module and the local 3d context encoding, respectively. Best scores are in \bf{bold}.}
\vspace{-2mm}
\label{tab:ablation}  
\end{table}

\begin{table}
\setlength{\tabcolsep}{3pt}
\resizebox{\linewidth}{!}{
\centering
\begin{tabular}{@{}l|c|c|c|c|c|c|c|c@{}}
    \hline
    \multicolumn{1}{c|}{\multirow{2}{*}{\textbf{Variants}}}  & \multicolumn{2}{c|}{\textbf{CD (cm) $\downarrow$}} & \multicolumn{2}{c|}{\textbf{Occ L1} $\downarrow$} & \multicolumn{2}{c|}{\textbf{Norm Cosine} $\downarrow$} & \multicolumn{2}{c}{\textbf{Norm L2} $\downarrow$}\\
    \cline{2-9}
    & mean & median & mean & median & mean & median & mean & median\\
    \hline
    2 views & 0.870 & 0.753 & 2.909 & 2.474 & 0.121 & 0.114 & 0.407 & 0.392\\ 
    4 views & 0.367 & 0.316 & 1.538 & 1.323 & 0.089 & 0.083 & 0.350 & 0.337\\
    6 views & \bf{0.279} & \bf{0.245} & \bf{1.383} & \bf{1.215} & \bf{0.082} & \bf{0.079} & \bf{0.337} & \bf{0.327}\\
    \hline
\end{tabular}
}
% \vspace{0.5mm}
\caption{Ablation studies on using different number of views as input. Best scores are in \bf{bold}.}
\vspace{-5mm}
\label{tab:ablation_views}
\end{table}

%%%%%%%%%%%%%%%%%%%%%%%%%%%%
\subsection{Comparisons}

\begin{figure*}[ptb]
\begin{center}
\includegraphics[width=0.95\linewidth]{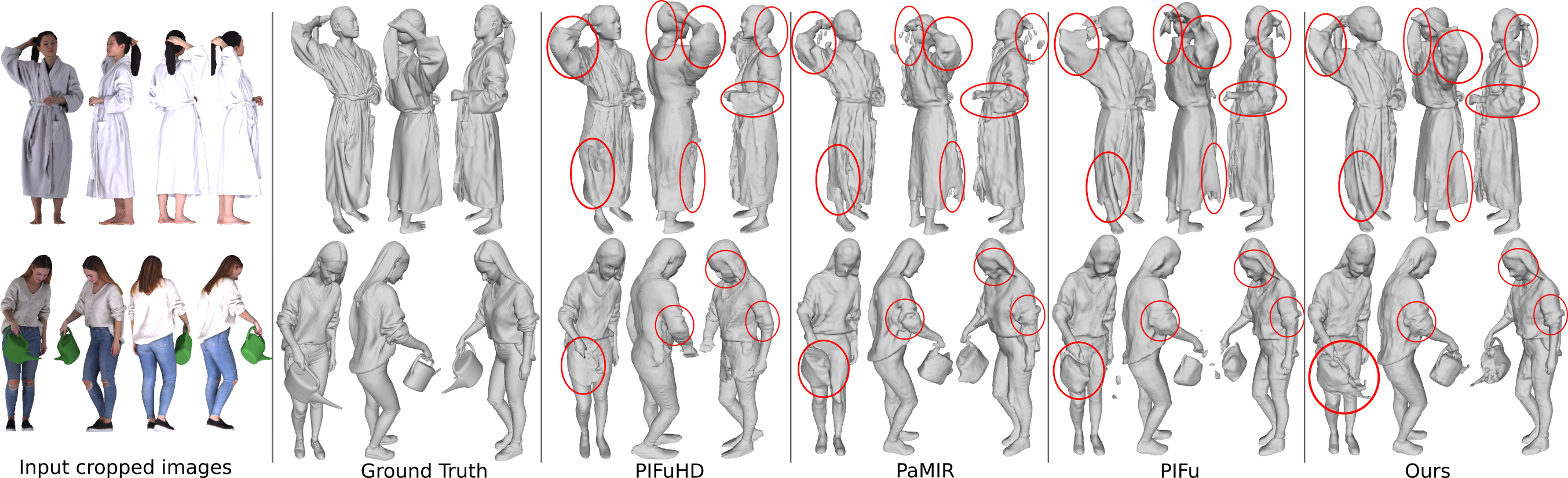}
\end{center}
\vspace{-3mm}
\caption{Qualitative results and comparisons with multi-view PIFu~\cite{pifuSHNMKL19}, multi-view PaMIR~\cite{DBLP:journals/corr/abs-2007-03858} and PIFuHD~\cite{DBLP:conf/cvpr/SaitoSSJ20}. The 4 input images are rendered with the rotations around the vertical axis : 10\degree, 110\degree, 150\degree, 300\degree. PIFuHD uses a single frontal view as input.}
\vspace{-3mm}
\label{fig:sota}
\end{figure*}

\begin{figure}[ptb]
\begin{center}
\includegraphics[width=0.9\linewidth]{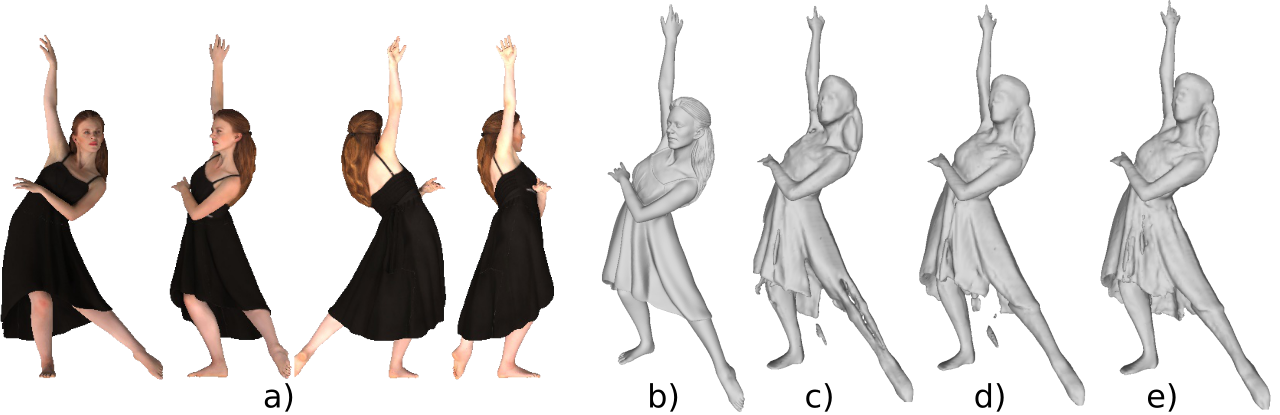}
\end{center}
\vspace{-3mm}
\caption{Ablation studies of our approach: a) Input cropped images. The 4 input images are rendered with the rotations around the vertical axis : 10\degree, 110\degree, 150\degree, 300\degree. b) Ground truth models. c) Ours without the attention-based view fusion module. d) Ours without the local 3D context encoding. e) Our full method.}
\label{fig:ablation}
\vspace{-3mm}
\end{figure}

\begin{figure}[ptb]
\begin{center}
\includegraphics[width=0.9\linewidth]{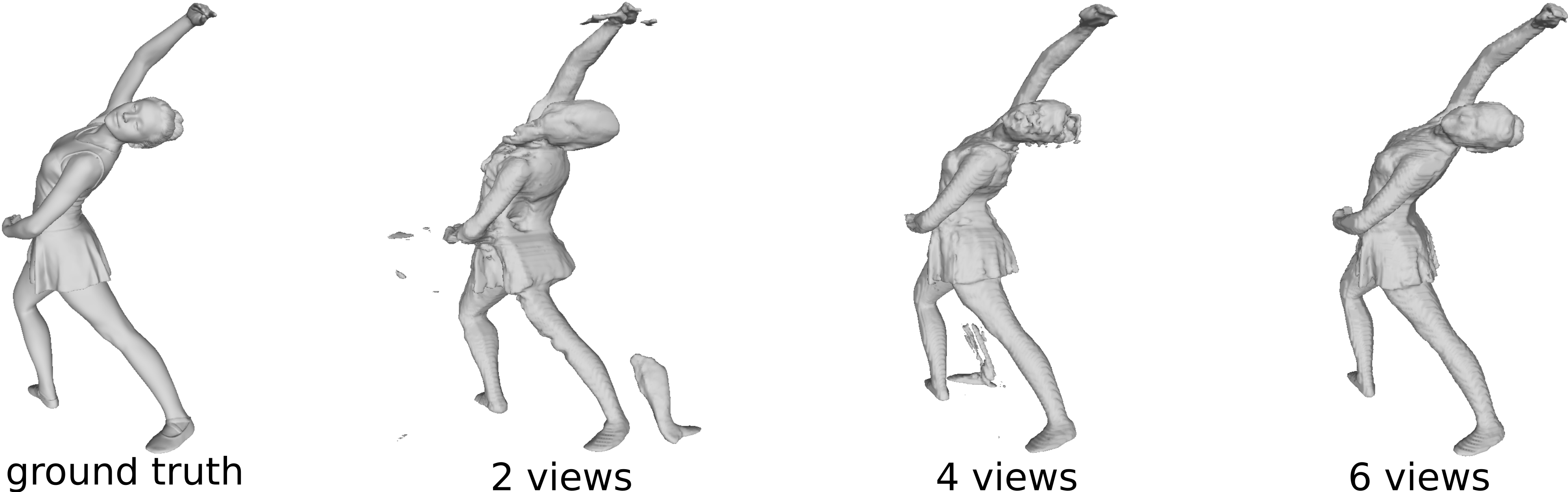}
\end{center}
\vspace{-3mm}
\caption{Ablation on different number of input views. As more views are added, the reconstruction with our method are improved.}
\vspace{-3mm}
\label{fig:ablation_views}
\end{figure}

In the context of 3D reconstruction of dressed humans from a few sparse views, PIFu~\cite{pifuSHNMKL19} demonstrated state-of-the-art results so we consider it as the baseline result. For the comparison we trained it on our training dataset. This method has proven its benefit against model-based reconstructions and we do not provide comparisons with the latter. PIFuHD~\cite{DBLP:conf/cvpr/SaitoSSJ20} extends PIFu to high resolution images and shows impressive single view reconstructions of details for the visible parts. No training code is available, so we use the published pre-trained model for the comparison. PAMIR~\cite{DBLP:journals/corr/abs-2007-03858} combines the implicit representation with a parametric body model and shows improved single-view and multi-view reconstructions. The released code and pre-trained model are only for single-view reconstruction, so we implemented the missing parts ourselves and trained a multi-view model on our training dataset. We do not provide direct comparisons between our method and multi-view stereo (MVS) methods applied on the exact same input data since MVS methods fail when only few images are available. 
PIFu, PIFuHD and PaMIR use orthographic images in which the human is at the center and cannot address the spatial consistency in world space. For a fair evaluation, we create a corresponding training / validation / test dataset composed of meshes from Renderpeople and evaluate all four methods on this data. 

Qualitative results on synthetic data are presented in Fig.~\ref{fig:sota}. PIFu and PaMIR achieve promising reconstructions but fail on some parts like the hair and the arm in the first row, or the watering can and clothing wrinkles in the second row. Our method appears clearly more robust and captures more geometric detail as can be seen on faces and clothing wrinkles.
PIFuHD achieves detailed reconstructions for the visible parts like the face but, unlike for our method, the quality decreases significantly for the hidden parts and the global shape is not respected like the head on both rows. This is inherent to single view reconstruction methods and emphasizes the utility of using multiple views.

This intuition is verified by the associated quantitative results in Tab.~\ref{tab:state_of_the_art} that confirm the benefit of our method on three aspects. First, the global quality of the reconstructions is improved by a large margin with the Chamfer distance. Second, metrics on surface normal are also in line and show that local geometric details are better captured. Third, our method achieves better results on the raw values of the implicit function.

%%%%%%%%%%%%%%%%%%%%%%%%%%%%
\subsection{Ablation Studies}

To evaluate the impact of our contributions, namely the multi-head self-attention fusion layer and the local 3D context encoding, we conducted qualitative and quantitative ablation studies. To isolate these contributions from eventual human center detection errors, we place here the human person at the center of the scene. For the first contribution, we replaced the view fusion module by a simple average pooling strategy and for the second, individual sample points were considered in place of the proposed local 3D grid.

Quantitatively, disabling the view fusion or the context encoding module both affect the reconstruction performance. From the results shown in Fig.~\ref{fig:ablation} and Tab.~\ref{tab:ablation}, we clearly see that the multi-head self-attention view fusion module is crucial for both the global quality and the local geometric details. On the other hand, the local 3D context encoding is not sufficient by itself but when combined with the view fusion module helps the global reconstruction quality and avoids holes or missing parts.

To evaluate the scalability of our method, we compare reconstructions with different numbers of input views. Visual results in Fig.~\ref{fig:ablation_views} show that the global quality of the shape (noise and missing parts) as well as geometric details (face and skirt) are improved as more views are used. Visual results are confirmed by the quantitative evaluation in Tab.~\ref{tab:ablation_views}. In particular, we observe a stronger improvement when using 4 views instead of 2 compared to 6 views instead of 4. This observation seems reasonable since the views used here are distributed evenly around the person and 4 views are sufficient to observe every side.
% The maximum number of views at training is related to the GPU memory usage. For example, 6 views is the limit for a Quadro RTX 5000 with 16Gb of memory. Using more views during training is possible at the cost of reducing the number of points sampled at each iteration, which can slow down training.

%%%%%%%%%%%%%%%%%%%%%%%%%%%%
\subsection{Spatially Consistent Reconstruction}

\begin{figure}[ptb]
\begin{center}
\includegraphics[width=0.85\linewidth]{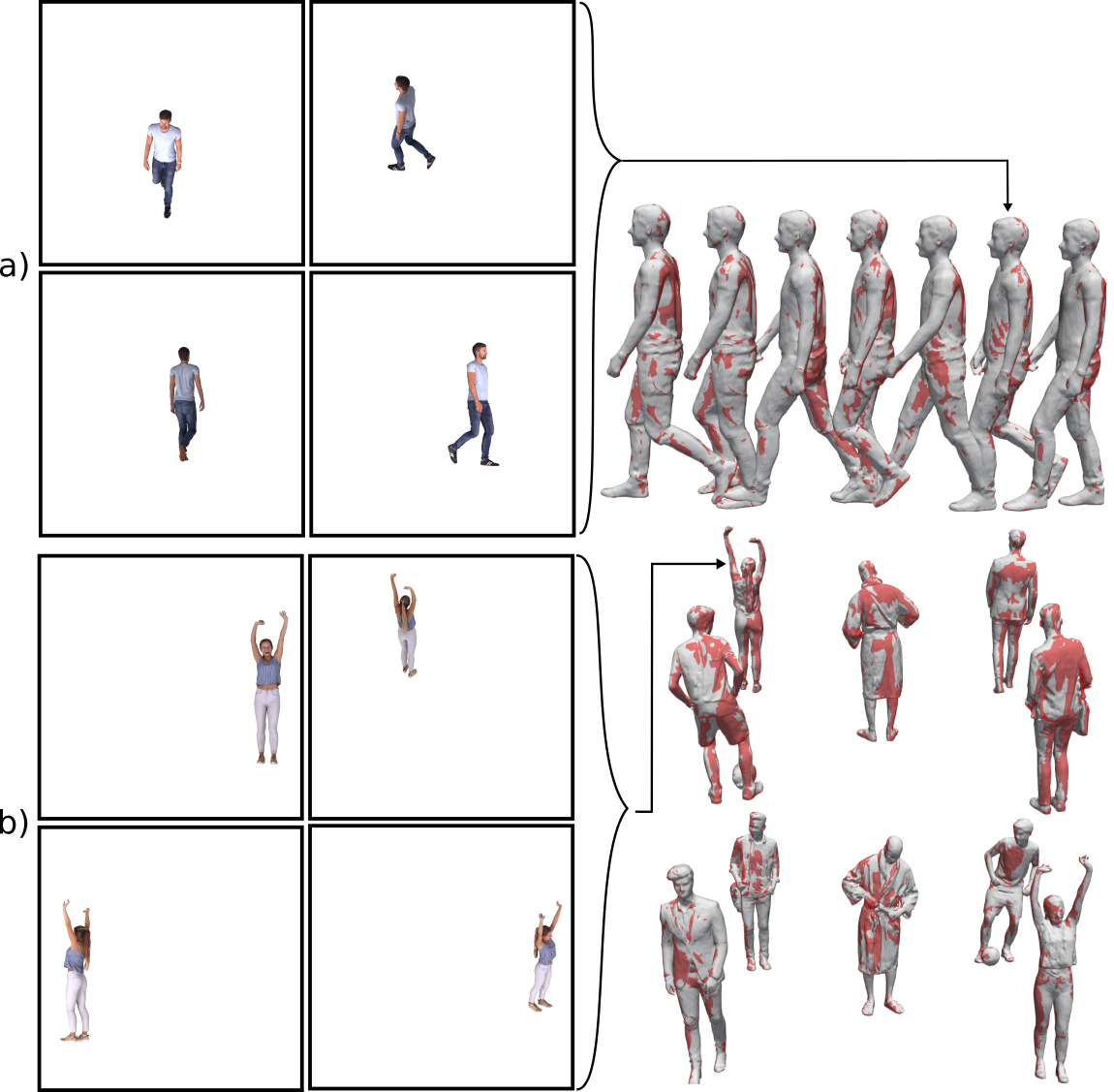}
\end{center}
\vspace{-4mm}
\caption{Spatially consistent reconstructions. a) Frame-by-frame reconstruction of a sequence from synchronized videos. Left: high resolution images for one example frame. Right: our result with the ground truth superimposed in red. b) Person-by-person reconstruction of a scene with multiple people. Left: high resolution images for one example person. Right:  our result with the ground truth superimposed in red. a) and b) For both, the camera rotations around the vertical axis are 10\degree, 110\degree, 200\degree and 300\degree with an random elevation angle between 0\degree and 40\degree.}
\vspace{-3mm}
\label{fig:spatial_consistency}
\end{figure}

To demonstrate the spatial consistency of the reconstructions we consider two scenarios, using data from Renderpeople~\cite{renderpeople}. First, we apply our method to dynamic input, namely to four synchronized video sequences showing a human walking in a scene. We reconstruct the sequence frame-by-frame, and Fig.~\ref{fig:spatial_consistency}(a) shows that the reconstructions contain details (ears, clothing wrinkles) and are spatially consistent with the ground truth. A better visualization is provided in the supplementary video.

As a second scenario, we consider a static scene containing multiple persons at different positions and render high resolution images with 4 cameras. Note that this evaluation focuses on spatially consistent reconstructions and not occlusions between persons. Hence, we render each person individually while the other persons are hidden. Fig.~\ref{fig:spatial_consistency}(b) shows that the reconstructions are spatially consistent with the ground truth and we can also note that the heights of the persons are correctly reconstructed.

\subsection{Application to Real-world Data}

To demonstrate the generalization of our method, we show 3D reconstructions of clothed humans with real images obtained with a $60$ camera multi-view capture system. We compare with PIFu and PIFuHD when reconstructing with the front view only, to PIFu when reconstructing with $4$ views, and to a multi-view stereo method~\cite{DBLP:conf/eccv/LeroyFB18} on the same scenes but with $60$ images. For all methods to be applicable, we consider the person centered in the middle of the scene. It is important to note that the networks were trained purely on synthetic data while tested on images from a real acquisition scenario. Fig.~\ref{fig:teaser} shows that single view reconstructions suffer from an inherent depth ambiguity: some parts are missing (hair and backpack) and the pose is incorrect. Our method performs better than PIFu when $4$ views are available, with more realistic global shapes and more detailed local geometries. More importantly, the comparisons with the multi-view stereo method applied to $60$ images demonstrate the potential of data-driven strategies in the multi-view reconstruction domain.

%%%%%%%%%%%%%%%%%%%%%%%%%%%%
\section{Conclusion}

In this paper, we build on recent progress on implicit representations of 3D data and propose a method for 3D reconstruction of clothed humans from a few sparse views. We introduce three key components: 1) a  spatially  consistent  reconstruction  that  allows  for  arbitrary  placement  of  the  person  in  the  input  views  using  a perspective camera mode; 2) a fusion layer based on an attention mechanism that learns to efficiently combine the information from all available views; 3) a mechanism that encodes local 3D patterns in the multi-view context. Our experiments show that the proposed method outperforms the state of the art in terms of details and global quality of the reconstructions on synthetic data. We also demonstrate a better generalization of our method on real data acquired with a multi-view platform. Additionally, we show that our approach can approximate multi-view stereo results with dramatically less views.

\section{Acknowledgements}
We thank Laurence Boissieux and Julien Pansiot from the Kinovis platform at Inria Grenoble and our volunteer subjects for help with the 3D data acquisition.

{\small
\bibliographystyle{ieee_fullname}
\bibliography{egbib}
}

\clearpage
\begin{center}
\LARGE{\textbf{Supplementary Materials}}
\end{center}
\normalsize
\vspace{2mm}

In the following, we first provide more details about our implementation. We give then additional qualitative results and ablation tests. Note that a supplementary video is also provided to better visualize the reconstruction results.

\thispagestyle{empty}

\section{Implementation}
\subsection{Training Views}
\label{view_selection}
To train our deep neural network, we created a synthetic model view set by rendering 3D models from Renderpeople~\cite{renderpeople} using $360$ cameras located around them as explained below. In contrast to Multi-View Stereo methods,  only a few of these views are considered at inference (between $2$ and $6$ in our experiments). 
The views used at inference should be ideally evenly distributed  around the person in order to increase its visibility. At inference, results are most of the time better for parts of the surface that are observed than  hidden ones for which the reconstruction relies solely on the prior learned from the training set. To build such image sets for the training we sample the synthetic views of a 3D model and create several model view subsets with few images. 

%During training, similar data should be given to the network, so at each training iteration it receives a few new view samples from the view set that are distributed around the person.

\begin{figure}[ptb]
\begin{center}
\includegraphics[width=1.0\linewidth]{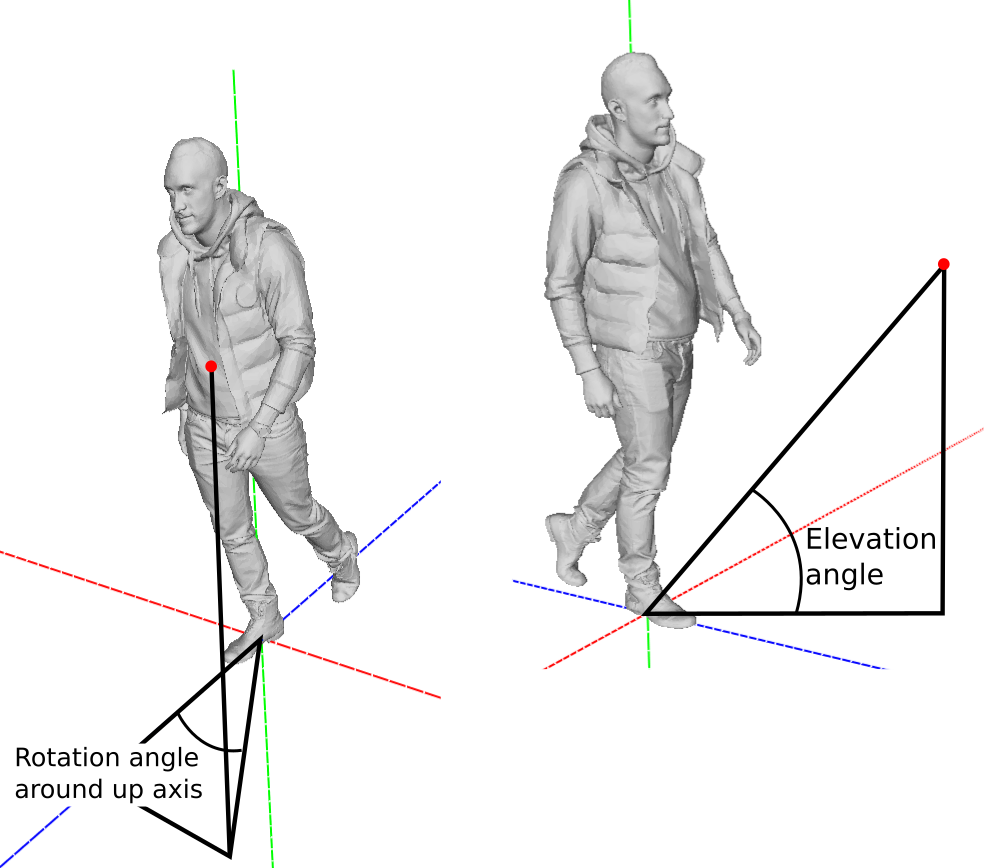}
\end{center}
\vspace{-4mm}
\caption{View selection angles.}
\vspace{-3mm}
\label{fig:angles}
\end{figure}

To define the position of our cameras when creating such a subset, we use a rotation angle around the up-axis and an elevation angle, as described in Figure~\ref{fig:angles}. For the orientation, we assume that the cameras are always looking at the center of the scene. 

In practice, at each training iteration we choose $N$ angles around the up axis that are evenly distributed among $\left[0\degree, 45\degree, 90\degree, 135\degree, 180\degree, 225\degree, 270\degree, 315\degree  \right] $ and add a random offset between $-20\degree$ and $20\degree$. The elevation angles are selected randomly between $0\degree$ and $45\degree$. Note that we trained our model with a fixed elevation angle when comparing with other methods (\ie PIFu~\cite{pifuSHNMKL19} and PIFuHD\cite{DBLP:conf/cvpr/SaitoSSJ20}) that  consider a similar scenario.

\subsection{Human Center Localization}

\begin{figure*}[ptb]
\begin{center}
\includegraphics[width=1.0\linewidth]{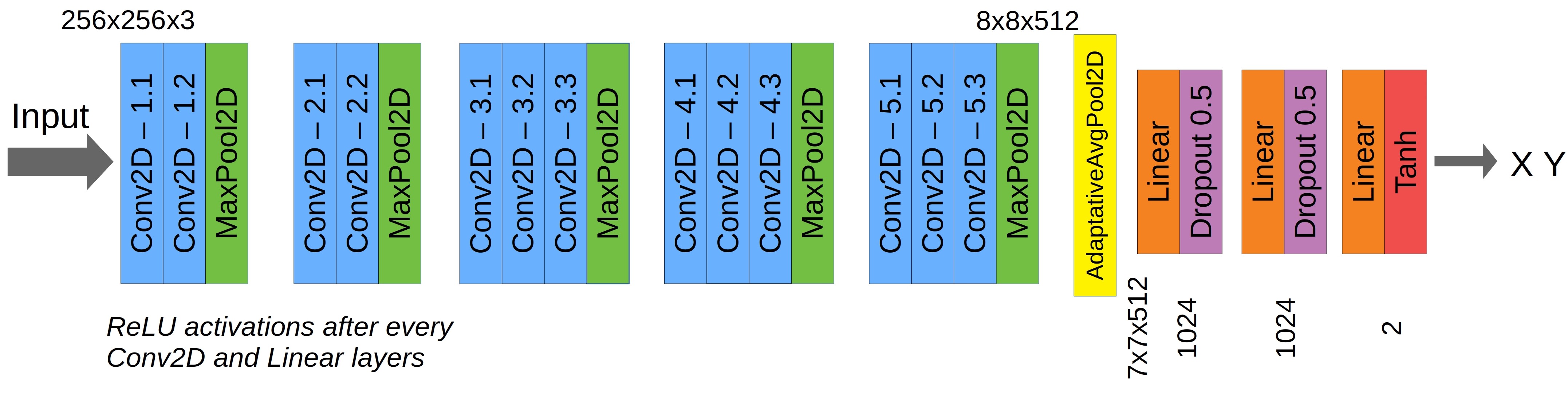}
\end{center}
\vspace{-4mm}
\caption{Human Center Detection network based on VGG16~\cite{DBLP:journals/corr/SimonyanZ14a}.}
\vspace{-3mm}
\label{fig:vgg16}
\end{figure*}

In Figure~\ref{fig:vgg16} we show the architecture of our deep neural network based on VGG16~\cite{DBLP:journals/corr/SimonyanZ14a} to detect the human center on each of the view. These 2D detections are then used to triangulate the 3D position of the person in the scene. The center of the person is arbitrarily defined but should be coherent with the origin of the canonical coordinate systems used at training. In practice, we defined it as : 
\begin{equation*}
	\begin{bmatrix} 
		median(vertices.x) \\ 
		 0.5 * (max(vertices.y) - min(vertices.y)) \\ 
		median(vertices.z)
	\end{bmatrix} 
% \left[ median(vertices.x), 0.5 * (max(vertices.y) - min(vertices.y)), median(vertices.z)  \right]
\end{equation*}
where $y$ is the up-axis. We do not use the median for the up-axis to account for cases where numerous vertices are grouped at the top or the bottom. Such cases are worth considering since a human is less symmetric with respect to the horizontal plane. In Table~\ref{tab:hcd}, we compute the $L_2$ distance between the 2D detections and 2D ground truth as well as the 3D positions triangulated from the 2D detection and the 3D ground truth. Here we used $4$ views evenly distributed around the person with a random elevation axis between $0\degree$ and $45\degree$.

\begin{figure}[ptb]
\begin{center}
\includegraphics[width=1.0\linewidth]{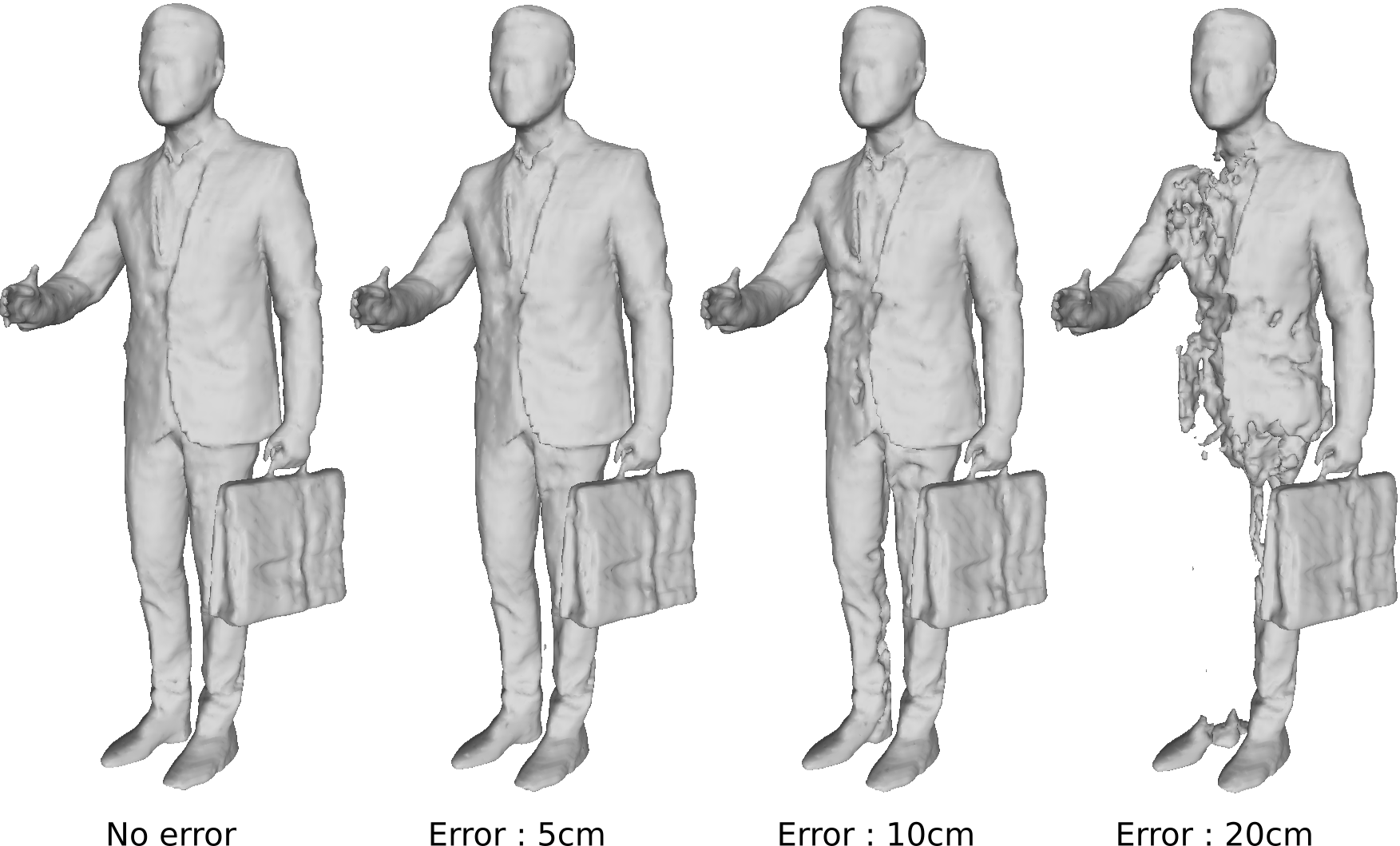}
\end{center}
\vspace{-4mm}
\caption{Reconstructions from 4 views that show the impact of a 3D human center localization error on the reconstruction.}
\vspace{-3mm}
\label{fig:impact_triangulation_error}
\end{figure}

As shown in Table~\ref{tab:hcd}, the average Euclidean distance between the ground truth and triangulated 3D human center is around $4.4 cm$. 
We compute these metrics on test data ($360$ groups of $4$ views for $50$ persons) and follows the strategy explained in Section~\ref{view_selection} to select the 4 views.
Additionally, we show in Figure~\ref{fig:impact_triangulation_error} an example of reconstruction with manually specified errors on the human center location. We see that the reconstruction quality is not affected too much up to 5cm. Noise starts being visible with an error of 10 cm and the reconstruction fails with larger error like 20 cm.

\begin{table}
\setlength{\tabcolsep}{3pt}
\centering
\begin{tabular}{|c|c|c|c|}
    \hline
    \multicolumn{2}{|c|}{\textbf{L2 - 2D (pixels)}} & \multicolumn{2}{c|}{\textbf{L2 - 3D (cm)}}\\
    \hline
    mean & median & mean & median\\
    \hline
    9.795 & 8.944 & 4.398 & 4.291\\ 
    
    \hline
\end{tabular}
% \vspace{0.5mm}
\caption{Evaluation of the human center detection on images and the 3D triangulated position of the center. Both are evaluated on test images.}
\vspace{-2mm}
\label{tab:hcd}
\end{table}

\section{Attention scores}

% add figure attention

\begin{figure}[ptb]
\begin{center}
\includegraphics[width=0.9\linewidth]{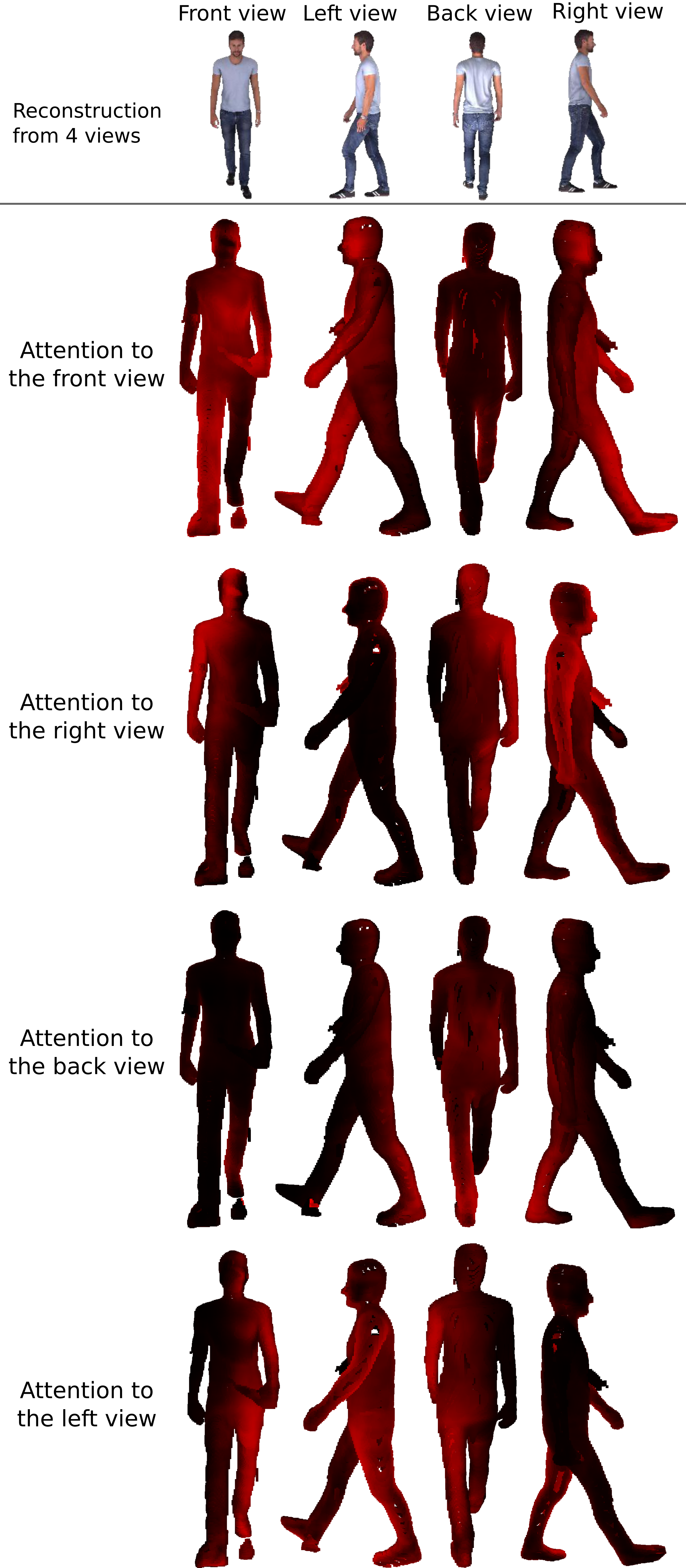}
\end{center}
\vspace{-4mm}
\caption{Attention scores: points predicted as close to the surface (in or out) are visualized. The intensity of the red channel represents the contribution of the considered view.}
\vspace{-3mm}
\label{fig:attention_scores}
\end{figure}

We provide in Figure~\ref{fig:attention_scores} a visualization of the attention scores of our view fusion module. We use 4 input views, evaluate our deep neural network in a 3D grid of resolution 256 and save the attention score of the first self-attention layer. Note that we use a single head for this experiment. Points that are predicted close to the surface inside or outside are visualized and the intensity of the red channel represents how much the considered view contributed for each point. We clearly see that each point attend more to views in which they are visible.

% Evaluation runtime ?

%%%%%%%%% BODY TEXT
\section{Additional visual results}

\subsection{Comparison with the state of the art}
\begin{figure*}[ptb]
\begin{center}
\includegraphics[width=1.0\linewidth]{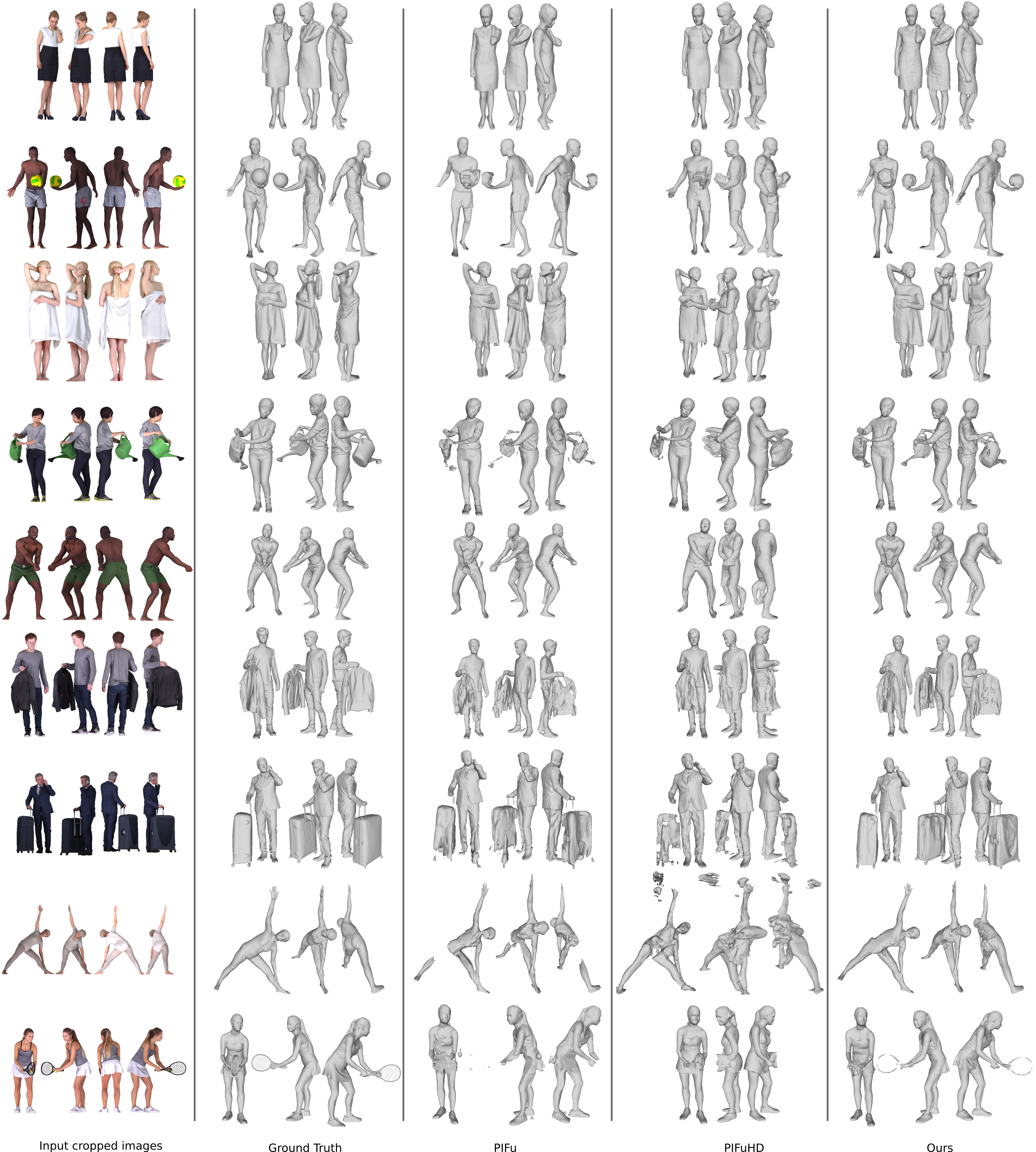}
\end{center}
\vspace{-4mm}
\caption{Qualitative results and comparisons. PIFu~\cite{pifuSHNMKL19} and our method take as input the 4 cropped images, whereas PIFuHD~\cite{DBLP:conf/cvpr/SaitoSSJ20} receives only the frontal view. The 4 input images are rendered with the rotations around the vertical axis : 10\degree, 110\degree, 150\degree, 300\degree.}
\vspace{-3mm}
\label{fig:sota_supp}
\end{figure*}

Qualitative visual comparisons between our proposed method, the considered baseline~\cite{pifuSHNMKL19} as well as state-of-the-art single-view reconstruction method PIFuHD~\cite{DBLP:conf/cvpr/SaitoSSJ20} are presented in Figure \ref{fig:sota_supp}. In particular, we note the improved global quality of the recovered accessories and the reduced level of noise in the reconstructions using our method. Moreover, sharper details on the faces and wrinkles on the clothes are recovered by our approach. Two difficult cases with less usual pose and thin structures are shown in the last two rows. Although our reconstructions contain some noise and missing parts, we can see a significant improvement compared to the other two methods.

\subsection{Application to real-world data}
\begin{figure*}[t!]
\begin{center}
\includegraphics[width=0.6\linewidth]{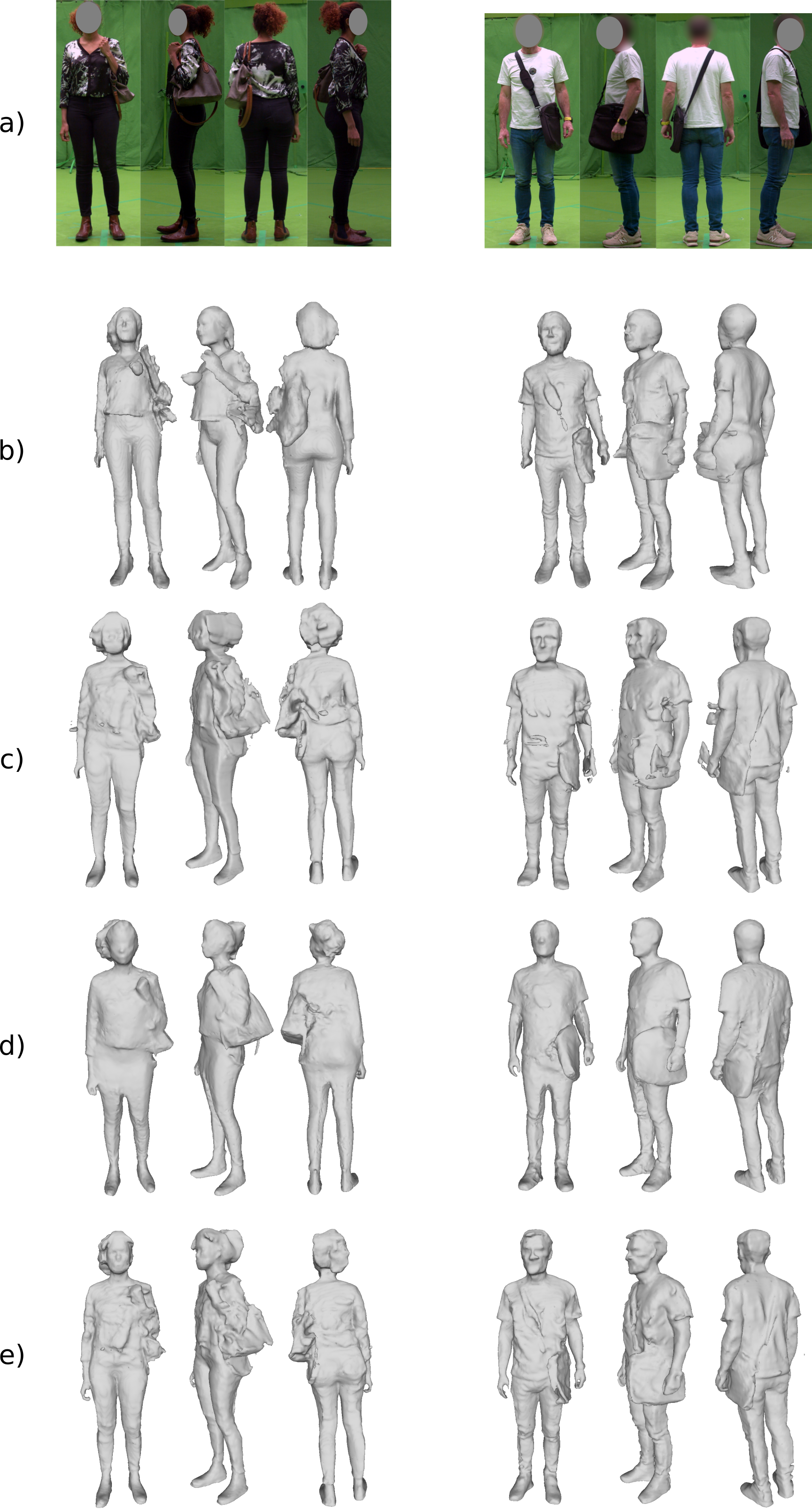}
\end{center}
\vspace{-4mm}
\caption{Qualitative results and comparisons with a real capture apparatus: a) real RGB images. b) single frontal view reconstruction using PIFuHD~\cite{DBLP:conf/cvpr/SaitoSSJ20}. c) 4-view reconstructions using PIFu~\cite{pifuSHNMKL19}.  d) 60-view reconstructions using a multi-view stereo approach~\cite{DBLP:conf/eccv/LeroyFB18}. e) 4-view reconstructions using our method.}
\label{fig:kinovis}
\vspace{-3mm}
\end{figure*}

A crucial aspect of our work is the applicability to real-world data. In Figure~\ref{fig:kinovis} we provide additional comparisons between our method, the considered baseline~\cite{pifuSHNMKL19}, state-of-the-art single-view reconstruction method~\cite{DBLP:conf/cvpr/SaitoSSJ20}, and a 60-view reconstruction obtained with a Multi-view stereo strategy~\cite{DBLP:conf/eccv/LeroyFB18}. In this real context, we observe that our method behaves better than the baseline and single-view reconstruction, especially with complex scenes, \eg with accessories. The improvement is less obvious, yet there, with persons in standard poses and without accessories (\eg columns $3$ in Figure~\ref{fig:kinovis}). In this case, the strategy from ~\cite{pifuSHNMKL19} already provides good results. Another interesting comparison in this figure is with multi-view stereo (row b). While the MVS strategy provides robust and accurate estimations of the global shapes, our data-driven strategy yields more local details.

\subsection{Ablation visual results}

\begin{figure*}[ptb]
\begin{center}
\includegraphics[width=1.0\linewidth]{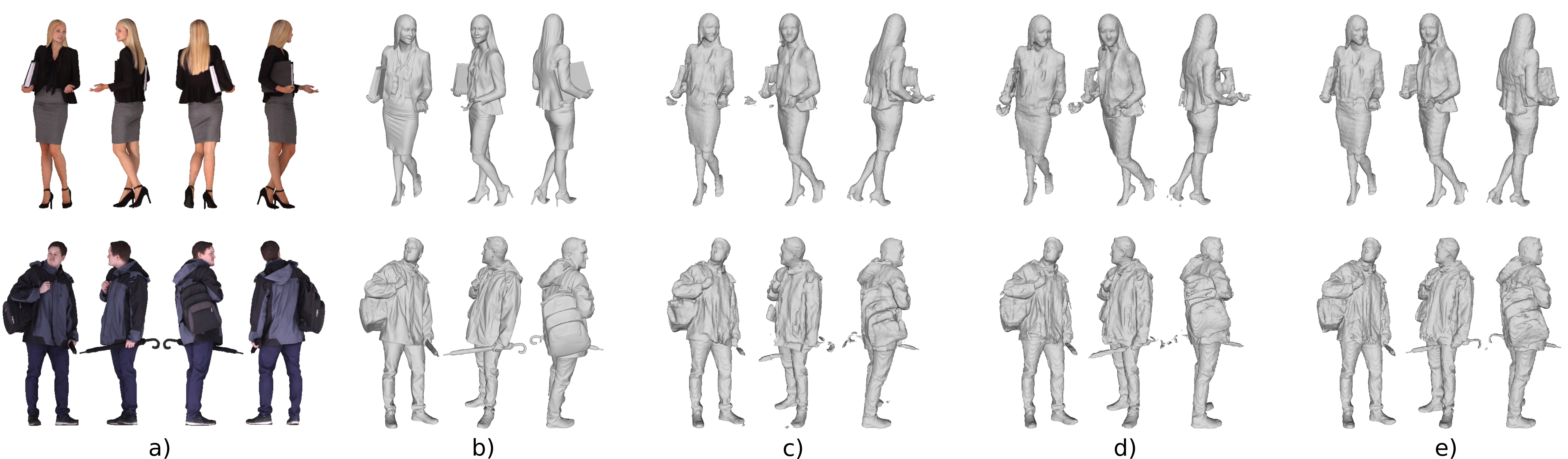}
\end{center}
\vspace{-4mm}
\caption{Ablation studies of our approach: a) Input cropped images. The 4 input images are rendered with the rotations around the vertical axis : 0\degree, 90\degree, 180\degree, 270\degree. b) Ground truth models. c) Ours without the attention-based view fusion module. d) Ours without the local 3D context encoding. e) Our full method.}
\label{fig:ablation_supp}
\vspace{-3mm}
\end{figure*}

Here, we show additional visual results of our ablation to evaluate the impact of our contributions. Quantitatively, disabling the view fusion or the context encoding module both affects the reconstruction performance. From the results shown in Fig.~\ref{fig:ablation_supp}, we clearly see that the multi-head self-attention view fusion module is crucial for both the global quality and the local geometric details. On the other hand, the local 3D context encoding impacts more the global quality of the reconstruction and helps avoiding holes or missing parts.

\subsection{Number of input views}

\begin{figure*}[ptb]
\begin{center}
\includegraphics[width=1.0\linewidth]{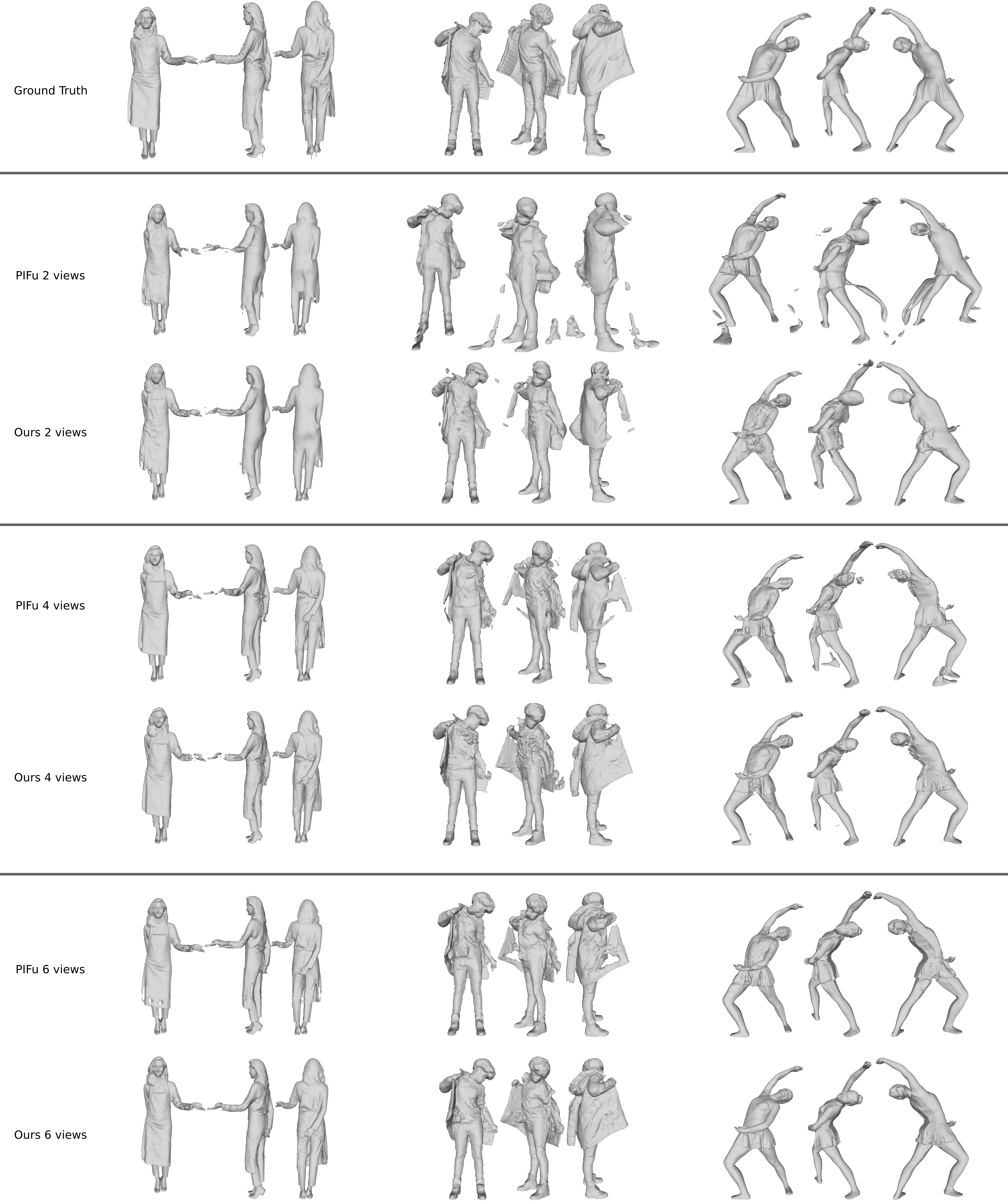}
\end{center}
\vspace{-4mm}
\caption{Impact of the number of input views on the reconstruction quality and comparison with PIFu~\cite{pifuSHNMKL19}.}
\vspace{-3mm}
\label{fig:ablation_views_supp}
\end{figure*}

Figure~\ref{fig:ablation_views_supp} shows results of our method with 2, 4 and 6 views as input. It demonstrates that adding views effectively decreases depth ambiguities and occlusions with a clear improvement in the reconstructions. It also shows the superiority of our proposed method compared to the baseline.

\begin{table}
\setlength{\tabcolsep}{3pt}
\resizebox{\linewidth}{!}{
\centering
\begin{tabular}{@{}l|c|c|c|c|c|c|c|c@{}}
    \hline
    \multicolumn{1}{c|}{\multirow{2}{*}{\textbf{Variants}}}  & \multicolumn{2}{c|}{\textbf{CD (cm) $\downarrow$}} & \multicolumn{2}{c|}{\textbf{Occ L1} $\downarrow$} & \multicolumn{2}{c|}{\textbf{Norm Cosine} $\downarrow$} & \multicolumn{2}{c}{\textbf{Norm L2} $\downarrow$}\\
    \cline{2-9}
    & mean & median & mean & median & mean & median & mean & median\\
    \hline
    PIFu 2 v. & 1.386 & 1.233 & 3.206 & 2.861 & 0.136 & 0.130 & 0.444 & 0.432\\ 
    PIFu 4 v. & 0.592 & 0.510 & 2.079 & 1.773 & 0.103 & 0.093 & 0.376 & 0.358\\
    PIFu 6 v. & 0.331 & 0.313 & 1.499 & 1.402 & 0.088 & 0.083 & 0.345 & 0.331\\
    \hline
    Ours 2 v. & 0.870 & 0.753 & 2.909 & 2.474 & 0.121 & 0.114 & 0.407 & 0.392\\ 
    Ours 4 v. & 0.367 & 0.316 & 1.538 & 1.323 & 0.089 & 0.083 & 0.350 & 0.337\\
    Ours 6 v. & \bf{0.279} & \bf{0.245} & \bf{1.383} & \bf{1.215} & \bf{0.082} & \bf{0.079} & \bf{0.337} & \bf{0.327}\\
    
    \hline
\end{tabular}
}
% \vspace{0.5mm}
\caption{Ablation studies on using different number of views as input. Best scores are in \bf{bold}.}
\vspace{-2mm}
\label{tab:nb_views}
\end{table}

\section{Additional ablations}

\subsection{Encoders}
\label{ssec_encoders}

\begin{figure*}[ptb]
\begin{center}
\includegraphics[width=1.0\linewidth]{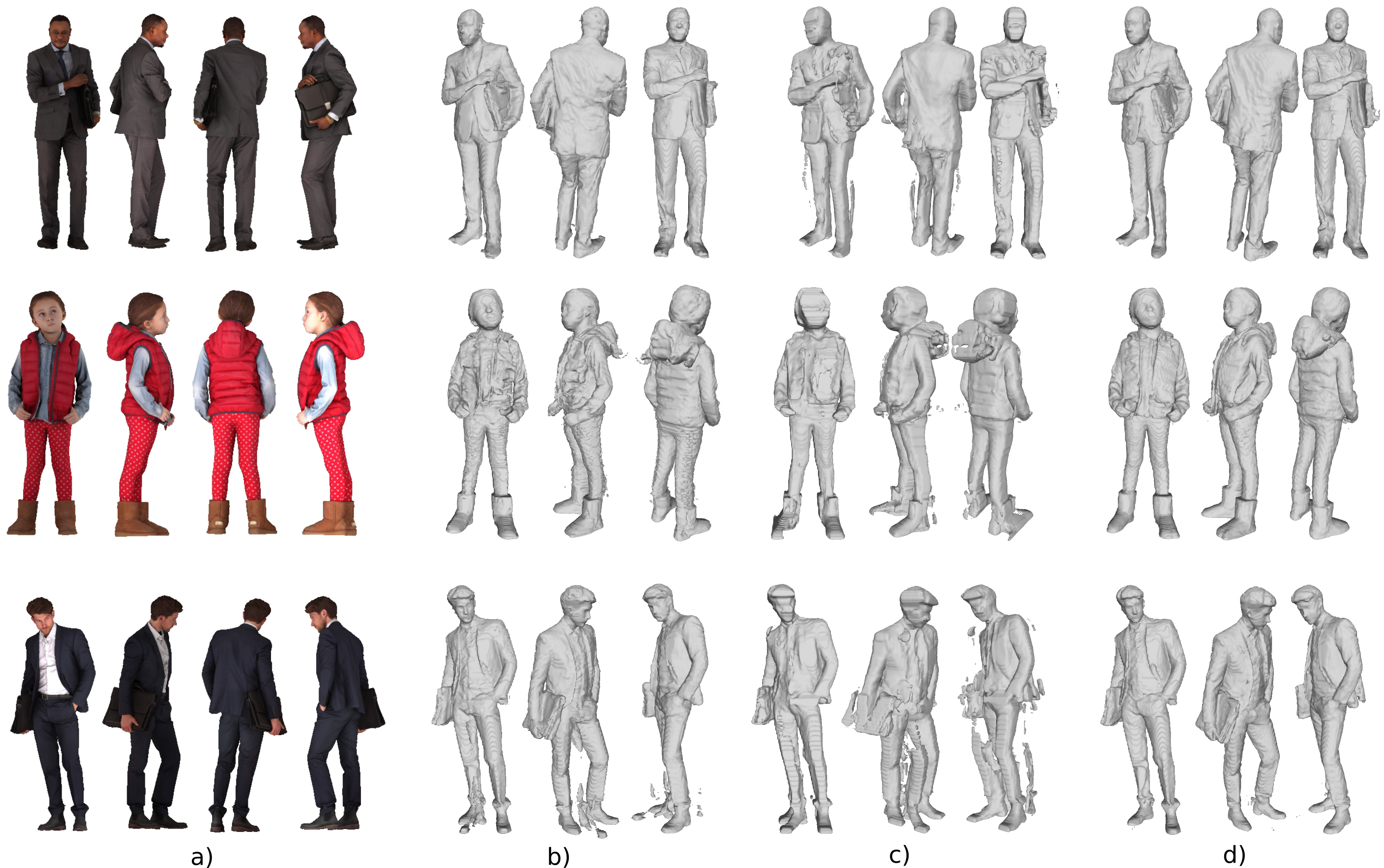}
\end{center}
\vspace{-4mm}
\caption{Comparative reconstruction results with our approach applied using $3$ different image encoders. a) Input RGB images. b) U-Net~\cite{ronneberger-2015}. c) HRNet~\cite{DBLP:journals/corr/abs-1910-05901}. d) Stacked Hourglass~\cite{newell-2016}.}
\vspace{-3mm}
\label{fig:encoders}
\end{figure*}

In our work, 2D features are extracted using the Stacked Hourglass encoder~\cite{newell-2016} that stacks multiple pooling and up-sampling networks. It allows the extraction of information at multiple scales and accounts therefore for both local and global contexts. Intermediate supervision is also applied to the output of each module while training our network. Of course numerous alternative encoders exist and could be used in our architecture in place of the Stacked Hourglass encoder. We provide in this section a comparison with $2$ popular options: U-Net~\cite{ronneberger-2015} and HRNet~\cite{DBLP:journals/corr/abs-1910-05901}. Results are shown in Figure~\ref{fig:encoders} and in Table~\ref{tab:encoders}. The U-Net~\cite{ronneberger-2015}, a fully convolutional network based on a contractive and an expansive part, gives results which are visually close to those obtained with the Stacked Hourglass encoder, with however significantly more noise as confirmed by the metrics in Table~\ref{tab:encoders}. On the other hand, the more recent work HRNet~\cite{DBLP:journals/corr/abs-1910-05901} fails to provide similar results in this context. 

\begin{table}[h]
\setlength{\tabcolsep}{3pt}
\resizebox{\linewidth}{!}{
\centering
\begin{tabular}{@{}l|c|c|c|c|c|c|c|c@{}}
    \hline
    \multicolumn{1}{c|}{\multirow{2}{*}{\textbf{Encoders}}}  & \multicolumn{2}{c|}{\textbf{CD (cm) $\downarrow$}} & \multicolumn{2}{c|}{\textbf{OCC L1} $\downarrow$} & \multicolumn{2}{c|}{\textbf{Norm Cosine} $\downarrow$} & \multicolumn{2}{c}{\textbf{Norm L2} $\downarrow$}\\
    \cline{2-9}
    & mean & median & mean & median & mean & median & mean & median\\
    \hline
    SHG & \textbf{0.385} & \textbf{0.322} & \textbf{1.602} & \textbf{1.380} & \textbf{0.087} & \textbf{0.081} & \textbf{0.343} & \textbf{0.326}\\
    U-Net~\cite{ronneberger-2015} & 0.572 & 0.482 & 1.984 & 1.688 & 0.108  & 0.101 & 0.389 & 0.369\\
    HRNet~\cite{DBLP:journals/corr/abs-1910-05901} & 1.092 & 1.075 & 3.682 & 3.547 & 0.181  & 0.178 & 0.565 & 0.553\\
    \hline
\end{tabular}
}
% \vspace{0.5mm}
\caption{Quantitative results obtained by our approach, on Renderpeople data~\cite{renderpeople}, with $3$ different image encoders (see text in Sec. \ref{ssec_encoders} for comments). Best scores are in bold.}
\vspace{-2mm}
\label{tab:encoders}
\end{table}

\subsection{Local grid size}

A key point of our method is the encoding of the local context of each sampled 3D point. To this purpose, we use a local 3D grid around each sampled point and in the pipeline, each original sampled point is associated with the additional points from their 3D local neighboorhood. At each training iteration, the local grids are aligned randomly with one of the camera used and the grid size is constant and defined before training.

% The local grid is defined as follows :

% \begin{equation}\label{eq:local_grid}
% \begin{aligned}
% & P = \{c + s . t . n_i \} \text{, where} \\
% & t \in \left\{-\left\lfloor \frac{grid_{res}}{2} \right\rfloor, -\left\lfloor \frac{grid_{res}}{2} \right\rfloor + 1, ..., \left\lfloor \frac{grid_{res}}{2} \right\rfloor\right\} \\
% & s = \frac{grid_{size}}{grid_{res}}\\
% \end{aligned}
% \end{equation}
% where $P$ represents points in the grid, $c \in \mathbb{R}^3$ is the local grid center and $n_i \in \mathbb{R}^3$ are the $27$ unit direct neighbors of the center.

Here we provide the results obtained with different grid sizes defined in world coordinates: small (2 cm), medium (10 cm) and large (20 cm) grids.

\begin{figure*}[ptb]
\begin{center}
\includegraphics[width=1.0\linewidth]{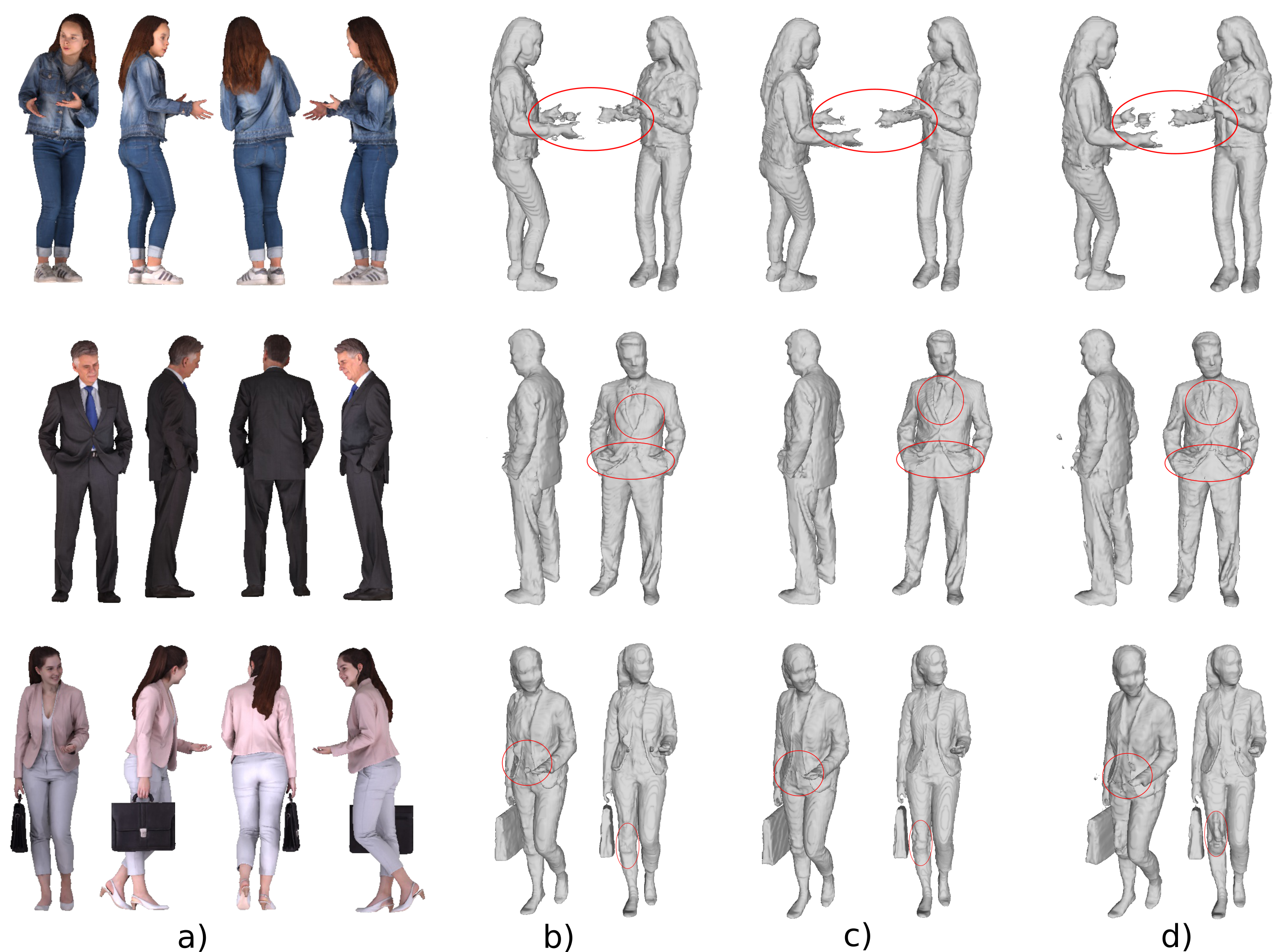}
\end{center}
\vspace{-4mm}
\caption{Qualitative comparison of the reconstructions using our method with 3 different local grid sizes. a) Input RGB images. b) Small grid (2 cm). c) Medium grid (10 cm). d) Large grid (20cm).}
\vspace{-3mm}
\label{fig:grid_size}
\end{figure*}

Table~\ref{tab:grid_size} shows that the best results were obtained with the medium-sized local grid, which is the one that was used for the other results in this paper. This result is confirmed visually on Figure~\ref{fig:grid_size}, where the medium grid shows better reconstructions with more details and less noise. This experiment demonstrates that the size of the local grid is important as it defines the neighborhood considered to predict the occupancy probability of the grid center. With a small grid, all grid points tend to be projected on the same 2D feature which prevents the 3D context to be encoded. On the other hand, with large grids, points can be far from each other, even on different body parts. In that case, the neighborhood considered is too large and not informative when predicting occupancies. 

\begin{table}[h]
\setlength{\tabcolsep}{3pt}
\resizebox{\linewidth}{!}{
\centering
\begin{tabular}{@{}l|c|c|c|c|c|c|c|c@{}}
    \hline
    \multicolumn{1}{c|}{\multirow{2}{*}{\textbf{Grid size}}}  & \multicolumn{2}{c|}{\textbf{CD (cm) $\downarrow$}} & \multicolumn{2}{c|}{\textbf{OCC L1} $\downarrow$} & \multicolumn{2}{c|}{\textbf{Norm Cosine} $\downarrow$} & \multicolumn{2}{c}{\textbf{Norm L2} $\downarrow$}\\
    \cline{2-9}
    & mean & median & mean & median & mean & median & mean & median\\
    \hline
    small (2 cm) & 0.422 & 0.413 & 1.668 & 1.566 & 0.089  & 0.087 & \textbf{0.342} & 0.336\\
    medium (10 cm) & \textbf{0.385} & \textbf{0.322} & \textbf{1.602} & \textbf{1.380} & \textbf{0.087} & 
    \textbf{0.081} & 0.343 & \textbf{0.326}\\
    large (20 cm) & 0.441 & 0.421 & 1.677 & 1.592 & 0.091  & 0.089 & 0.350 & 0.341\\
    
    \hline
\end{tabular}
}
% \vspace{0.5mm}
\caption{Quantitative results and comparisons with $3$ local grid sizes on Renderpeople data~\cite{renderpeople}. Best scores are in bold.}
\vspace{-2mm}
\label{tab:grid_size}
\end{table}

\clearpage

\end{document}